\documentclass[final,5p,times,authoryear]{elsarticle}
\usepackage{amssymb}
\usepackage{amsmath}
\usepackage{lineno}
\usepackage{xcolor}
\usepackage{hyperref}
\usepackage{soul}
\usepackage[percent]{overpic}
\usepackage{setspace}
\journal{Scientific Reports}

\begin{document}
\newcommand{\xb}{\mathbf{x}}
\newcommand{\rt}[1]{{\color{blue}{RT: #1}}}
\newcommand{\mm}[1]{{\color{magenta}{MM: #1}}}

\newcommand{\modified}[1]{#1}

\begin{frontmatter}

\title{Physics-informed convolutional neural networks for fluid flow through porous media}

\author{Rafał Topolnicki}
\affiliation{organization={Dioscuri Center in Topological Data Analysis, Institute of Mathematics, Polish Academy of Sciences},addressline={ul. Śniadeckich 8},postcode={00-656},city={Warsaw},country={Poland}}
\affiliation{organization={Institute of Experimental Physics, Faculty of Physics and Astronomy, University of Wrocław}, addressline={pl. M. Borna 9},postcode={50-204},city={Wrocław},country={Poland}}

\author{Paweł Dłotko}
\affiliation{organization={Dioscuri Center in Topological Data Analysis, Institute of Mathematics, Polish Academy of Sciences},addressline={ul. Śniadeckich 8},postcode={00-656},city={Warsaw},country={Poland}}

\author{Maciej Matyka*}
\affiliation{organization={Institute of Theoretical Physics, Faculty of Physics and Astronomy, University of Wrocław},addressline={pl. M. Borna 9},city={Wrocław},postcode={50-204},country={Poland}}
\affiliation{organization={Parallel and Distributed Systems Laboratory, Jožef Stefan Institute},addressline={Jamova cesta 39},city={Ljubljana},postcode={1000},country={Slovenia}}
\affiliation{organization={*Corresponding author: maciej.matyka@uwr.edu.pl (Maciej Matyka)}}

\begin{abstract}
Accurate simulation of fluid flow in porous media is a challenging task due to the complexity of pore-space geometries and the computational cost of solving the Navier–Stokes equations. Traditional numerical solvers rely on carefully constructed meshes, often requiring manual intervention, and typically exhibit slow convergence. This difficulty is particularly pronounced in porous media, where the diffusive nature of momentum transport is hindered by intricate solid boundaries. These challenges limit the efficiency of numerical simulations, particularly when repeated evaluations are required. 
We present a neural-network-based framework for predicting pore-scale velocity fields directly from sample geometry. The method is based on a convolutional encoder–decoder architecture with skip connections, designed to preserve fine-grained structural information. Physical consistency is encouraged through a custom loss function composed of multiple terms: incompressibility, no-flow conditions within solids, periodicity constraints, and agreement with the global tortuosity index. We systematically analyze the influence of weight selection for these loss terms, quantifying their individual contributions to prediction accuracy.
Several architectural variants inspired by computer vision are evaluated to identify one providing the best performance and robustness. The generalization ability of the trained network is assessed on samples outside the training distribution, including variations in boundary conditions, obstacle geometry, and porosity.
Finally, we demonstrate additional practical applications in which network predictions are used to initialize the Lattice–Boltzmann simulations, a standard fluid dynamics solver, often used in complex boundary problems like porous media and used by us to train the network.  We have used network-generated velocity field as a starting point and found that this significantly accelerates LBM solver convergence, achieving improvements in over 90\% of cases.
\end{abstract}


\end{frontmatter}

\section{Introduction} \label{sec1}

Porous materials are ubiquitous in nature and technology. Research on them covers a wide range of topics related to fluid transport, mechanics, and strength. One area with particular applications is the study of the permeability of porous samples, which has applications in oil extraction \cite{Zhong21}, CO$_2$ storage \cite{Cossins23}, and in medicine, with a prominent example of the blood-brain barrier as the drug delivery path, of which permeability is often of interest \cite{Fong15}.

For the past few decades, solving the fluid transport problem required the use of advanced numerical methods to solve the Navier-Stokes equations \cite{Anderson2002}. The main time-demanding elements of standard computational fluid dynamics simulation pipelines are: solving large systems of linear equations, creating computational grids, utilizing high-powered computers, and the diffusive nature of momentum transport. In general, given boundary conditions, initial conditions, and physical parameters, the only way to find the trajectories of particles in water is to solve the non-linear Navier-Stokes equations in approximate, numerical, and iterative approaches. The use of a mesoscopic Lattice-Boltzmann Method \cite{Guo13}, in which we deal with the transport of a particle distribution function on regular grids, is promising and successful, particularly in irregular and porous geometries \cite{Succi2001}. It comes, however, at an additional physical complexity in interpretation and additional cost of memory and time demand, thus requiring special optimizations \cite{Lehmann22} and memory layouts for handling large datasets \cite{Tomczak19}.

Meanwhile, we (humans) learn by observing physical systems. Looking at the ball thrown to the sky, we know that it will eventually fall, and we do this naturally, not knowing the PDEs behind. Similarly, for fluid flow in an open channel with an obstacle, we know approximately where the water will go and turn, as well as under what conditions a vortex may form. It is intuitively consistent that vortex-dominated flow occurs under conditions of increased velocity, and we are able to predict based on our knowledge and previous observations. Recent research has shown that this is also possible using artificial intelligence and deep convolutional neural networks. Guo et al. have demonstrated steady-state flow predictions for the flow over single objects immersed in a channel, utilizing convolutional neural networks (CNNs) \cite{Guo16}. Recently, the same authors have shown unsteady flow predictions over single obstacles in a channel \cite{Guo24}. We are, however, aiming to predict the flow in complex geometries of porous media, both at low and high porosity, where the complexity of the pore network poses a challenge for neural networks.
In our recent work we have shown, that only geometry is enough to predict physical properties of porous samples to predict their macroscopic properties in fluid flow and diffusive processes using CNN's \cite{Graczyk20, Graczyk23}. 
Research record on the topic of predicting physical properties of physical samples using deep learning techniques is rich. Recently, Lin et al. studied the application of CNNs in predicting velocity and temperature distributions in metal foam samples, demonstrating the outstanding efficiency of the method compared to standard numerical procedures \cite{Lin25}. 
It has been shown that the prediction of physical fields using CNNs can be improved by incorporating a physically informed network to reproduce temperature fields \cite{Zhao23}, a procedure that can be equivalent to solving differential equations and serves to generate physically correct results.

Our work proposes using a combined physics-informed CNN approach for fluid flow in complex porous media at various porosity conditions. We use convolutional neural networks and a custom, physically motivated loss function. 
\modified{
The main contributions of our work within CNN-based pore-scale flow prediction are summarized as follows:}

\begin{itemize}
\item[1)] We implemented a complete process from constructing the porous samples, simulation using the Lattice Boltzmann Method, through CNN training for predicting velocity fields in fluid flow through a porous media at various porosities.

\item[2)] \modified{
We construct and evaluate a custom loss function tailored to pore-scale flow prediction, including terms related to periodicity of the computational grid and tortuosity matching.
}.

\item[3)] We carried out tests on various types of porous media samples to demonstrate the generalization abilities of the neural network and predict beyond the data region on which the network was learned. 

\item[4)] We show practical application of CNNs for the Lattice Boltzmann Method: LBM solver can be significantly accelerated if the neural network's predictions are used to initialize the fluid dynamics solver.

\end{itemize}

We note that the general use of CNNs as surrogate models and the idea of incorporating physics-inspired constraints into neural-network training are well established. The contribution of the present work is therefore not the introduction of a new learning paradigm, but rather the formulation and evaluation of a problem-specific physics-informed loss for pore-scale flow prediction, together with a systematic assessment of its components, architecture dependence, out-of-distribution behavior, and use for LBM initialization.

The paper is organized as follows: in Sec.~\ref{secMAT}, we introduce the physical model of porous media and discuss the fluid flow solver, the Lattice Boltzmann Method, which is used later to prepare the training velocity fields. We describe the procedure to synthesize porous medium models investigated in the paper. Additionally, the learning process is described in detail, including data augmentation and a detailed description of the loss function. We discuss the choice and analysis of the neural network architecture and describe details of the training protocol. In Sec.~\ref{secRES} we present the results of predictions made by the trained network, both for tortuosity and permeability of the samples. Then, we discuss and present results on the generalization properties of our approach, including its ability to generalize to other porosities, shapes of obstacles, and other boundary conditions. We summarize our findings in Sec.~\ref{secCON}.

\section{Materials and Methods}\label{secMAT}

\subsection*{Pore scale flow}

We describe porous samples at the pore-scale level, where solid obstacles, distributed randomly, fill the space up to the desired porosity of the sample. We assume periodic boundary conditions in both directions, which, for the two-dimensional case, directly represent a porous medium on a torus topology. The flow is driven by gravity, which accelerates the fluid to a certain level. Due to the no-slip boundary conditions, the system stabilizes to a so-called stationary flow. The governing equations for the fluid flow in pores are Navier-Stokes, that, for an incompressible fluid takes form:

\begin{equation}
\nabla \cdot \mathbf{u}=0,
\label{eq:nse1}
\end{equation}
\begin{equation}
\varrho(\mathbf{u} \cdot \nabla \mathbf{u})=-\nabla p+\mu \nabla^2 \mathbf{u}+\mathbf{f},
\label{eq:nse}
\end{equation}
where $\mathbf{u}$ is velocity, $\mu$ is dynamic viscosity, $p$ is the pressure, and $\varrho$ is its density.
Here, we operate in the regime of low Reynolds numbers, for which the flow through porous matrix satisfies simpler Darcy's linear law:
\begin{equation}
\mathbf{q}=-\frac{k}{\mu}\left( \nabla P + 
\varrho\mathbf{f}\right),
\end{equation}
where $q$ is Darcy flux, $k$ is permeability $[L^2]$ and $f$ is the external force density (i.e. gravity). 
\modified{
Permeability, $k$, is of utmost importance in applications and porous media research. It became a standard measure with a described protocol for the experimental treatment. For practical applications, it can be expressed in terms of tortuosity and porosity, i.e.:
\begin{equation}
k=\frac{\varphi^3}{c\tau^2S},
\end{equation}
where $c$ is a material-specific constant, $\tau$ is tortuosity of the medium, and $S$ is specific surface area of pores \cite{Koponen1997}. Such a definition of permeability allows for estimations based on the geometrical properties of the medium, except for the value of $\tau$, which must be measured in the simulation or experiment. Tortuosity is a non-dimensional number representing elongation of porous media pores due to existence of solid matrix and it is driven by pore space geometry and physical process of channel formation. It can be investigated using streamlines generated on top of the velocity field, whose averaged elongation can be used to compute the tortuosity index \cite{Koponen96,Matyka08}.
We will use the velocity field to compute tortuosity by directly integrating the streamwise and total momentum in the pore space, which results in the following expression for tortuosity:
\begin{equation}
    \tau=\frac{\langle v\rangle}{\langle v_x\rangle},
\end{equation}
}
where brackets denote average over pore space, and $x$ is the component of velocity parallel to the direction of external body force \cite{Duda11,Matyka12}. 

\subsection*{The Lattice Boltzmann Method}
To solve the flow problem and obtain pore-space velocity fields to train the neural network, we utilize an indirect, mesoscopic Lattice Boltzmann Method (LBM). LBM works on a regular grid with a nine-component velocity discretization. It solves the Navier-Stokes flow equations indirectly by solving the transport equation of the particle distribution function $f(\mathbf{x},t)$, which represents the probability of finding particles at a given velocity at a specific location and time. The method demonstrated its unique properties and accuracy in porous media flows, becoming popular primarily due to its simplicity in implementation and the ability to handle complex geometries using local no-slip boundary conditions.
We will utilize its properties and solve the transport equation of the particle distribution function using the following model in the BGK (one-relaxation approximation) collision term \cite{Succi2001}:
\begin{equation}
f_i(\mathbf{x}+\mathbf{e}_i,t+\delta t)=f_i(\mathbf{x},t)+\frac{f_i^{eq}(\mathbf{x},t)-f_i(\mathbf{x},t)}{\tau},
\end{equation}
where $\tau$ is relaxation time, and $f_i^{eq}$ is distribution function in equilibrium.
Calculation of consecutive moments of the transported distribution function gives the macroscopic density $\varrho(\mathbf{x},t)$ and velocity $\mathbf{u}(\mathbf{x},t)$ fields:
\begin{equation}
    \varrho(\mathbf{x},t)=\sum_i f_i(\mathbf{x},t),
\end{equation}
\begin{equation}
    \varrho(\mathbf{x},t)\mathbf{u}(\mathbf{x},t)=\sum_i f_i(\mathbf{x},t)\mathbf{e_i},
\end{equation}
where $\mathbf{e_i}$ are the lattice vectors.

\subsection*{Porous samples}

Two-dimensional porous structures were represented as binary images with a resolution of $256 \times 256$ pixels. The structures were generated by superimposing standing sinusoidal waves with a fixed wavelength and amplitude but varying random directions $\mathbf{q}_i$ and phases $\phi_i$. The resulting field is given by:  

\begin{equation}
f(\xb) = \sqrt{\frac{2}{N}} \sum_{i=1}^{N} \cos(\mathbf{q}_i \cdot \xb + \phi_i), \quad \xb \in [0,1] \times [0,1],
\label{eqn:random_trig}
\end{equation}  

where $\xb$ is the position vector, and the phases $\phi_i$ are randomly sampled from the interval $[0, \pi]$. The wave vectors are defined as $\mathbf{q}_{i,j} = 2k\pi$, where $k$ is a randomly chosen integer within the range $[-12,12]$. For each generated structure, the number of superimposed standing waves, $N$, is randomly selected beforehand from the range $[10, 100]$.  

Given the random function (\ref{eqn:random_trig}), the different phases of the system are defined as  
\begin{eqnarray} 
\xb \in \mathcal{B} & \textrm{if} & f(\xb) \le \varepsilon, \nonumber \\ 
\xb \in \mathcal{P} & \textrm{if} & f(\xb) > \varepsilon. 
\label{eqn:random_threshold}
\end{eqnarray}  
Here, $\mathcal{B}$ represents the solid phase, while $\mathcal{P}$ corresponds to the pore phase. The threshold value $\varepsilon$ is chosen to achieve the desired porosity $\varphi$ for each generated sample. The porosity values are uniformly sampled within the range $0.70 \leq \varphi \leq 0.95$.  

In this context, the porosity is defined as the ratio of void-phase pixels to the total number of pixels. Structures that do not allow fluid flow through the material were removed from the dataset. A total of 10,000 structures were generated and used for model training.  

\begin{figure}
    \centering
    \includegraphics[width=0.32\linewidth]{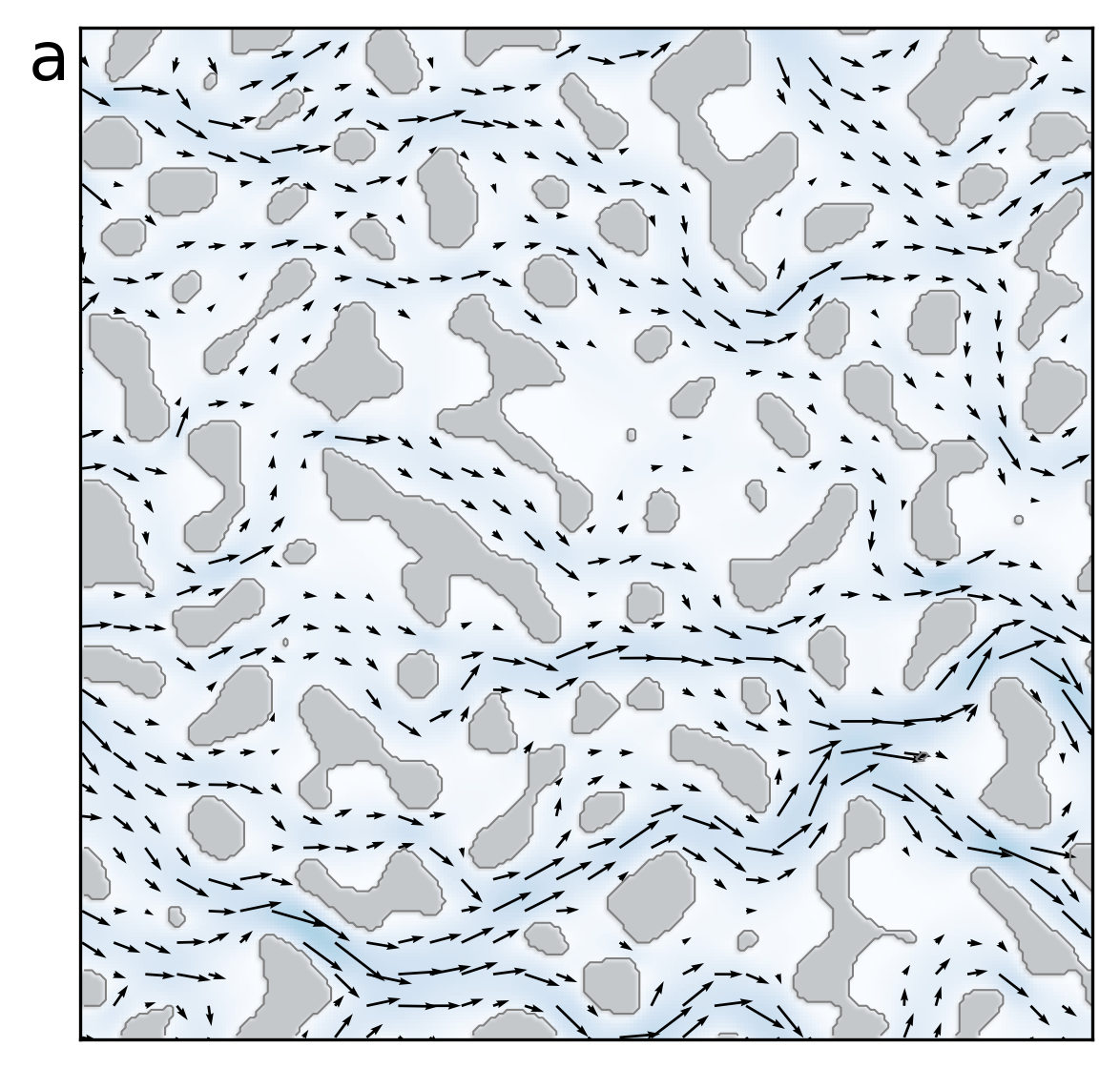}
    \includegraphics[width=0.32\linewidth]{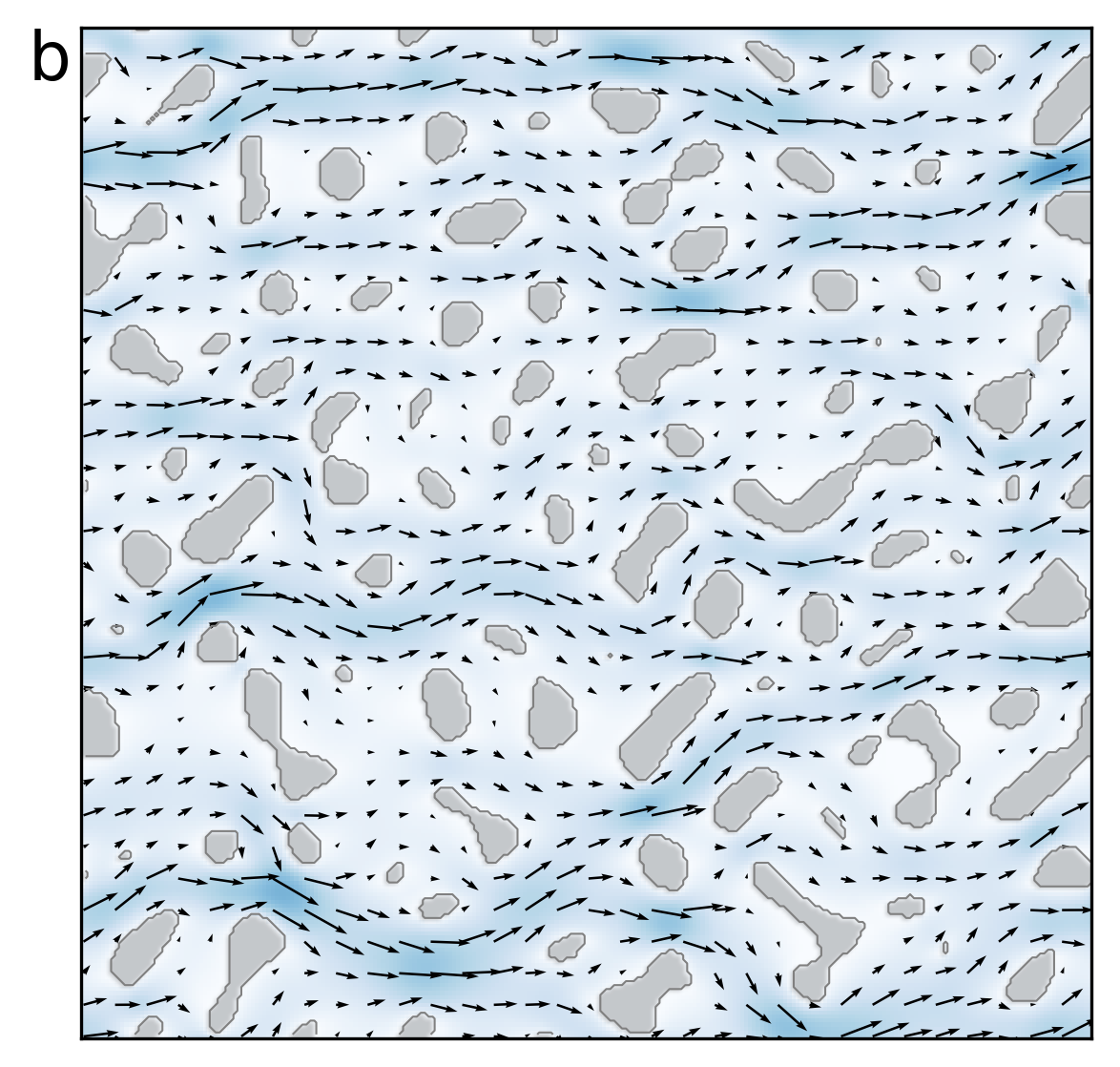}
    \includegraphics[width=0.32\linewidth]{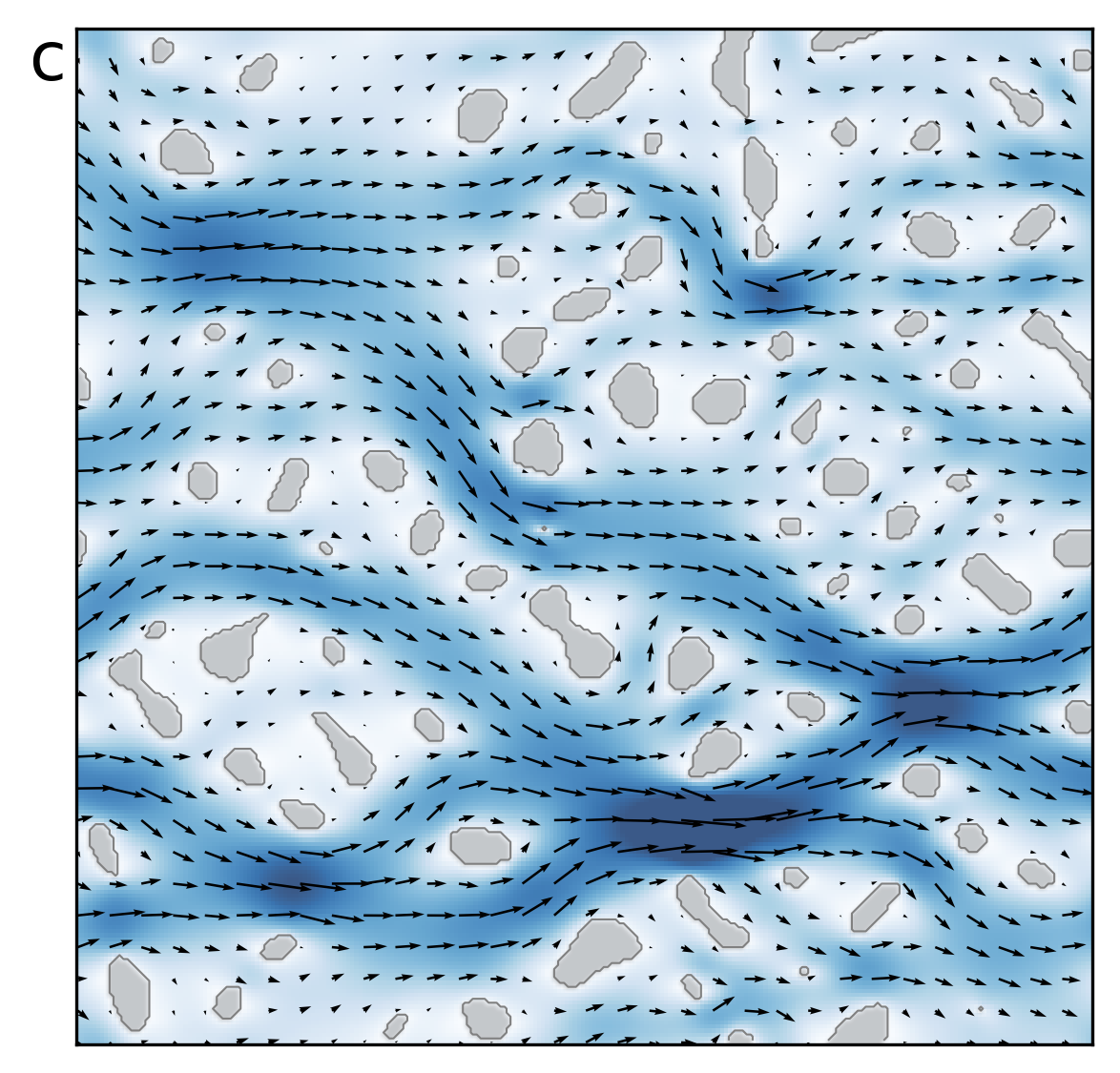}
    \caption{Example porous structures and corresponding LBM solutions - color indicates velocity magnitude. The same color scale applied on all plots. Structure porosity is (a) 0.771, (b) 0.861, (c) 0.917.  
    \label{fig:exampleporous}}
\end{figure}

Example of structures generated using this algorithm, along with its corresponding velocity fields, are presented in Figure~\ref{fig:exampleporous}.

\begin{figure*}
\centering
\includegraphics[width=\linewidth]{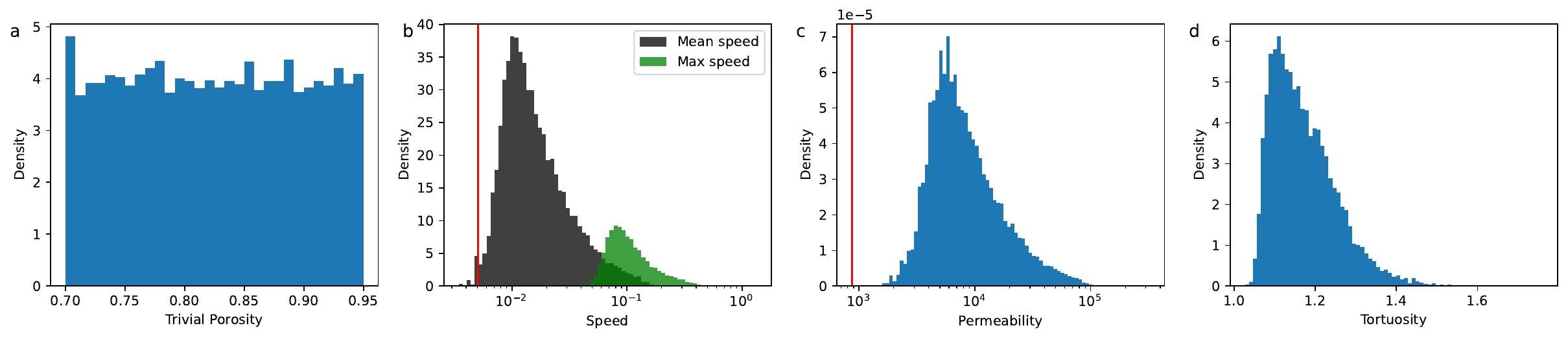}
\caption{
Statistical summaries of the LBM solutions for the dataset used to train the model.  
(a) Distribution of the trivial porosity (defined as the ratio of void area to total sample area).  
(b) Distribution of the mean and maximum fluid velocity within the porous structures.  
(c) Distribution of permeability values.  
(d) Distribution of tortuosity values across all samples. 
Red vertical lines on panels b) and c) indicate the prediction accuracy as indicated by the RMSE for the best model -- see Table \ref{tab:results}.
}
\label{fig:summaries}
\end{figure*}
Figure~\ref{fig:summaries} presents the distributions of key physical and geometric properties derived from the dataset used for model training.

\subsection*{Data augmentations}
\begin{figure}
    \centering
    \includegraphics[width=\linewidth]{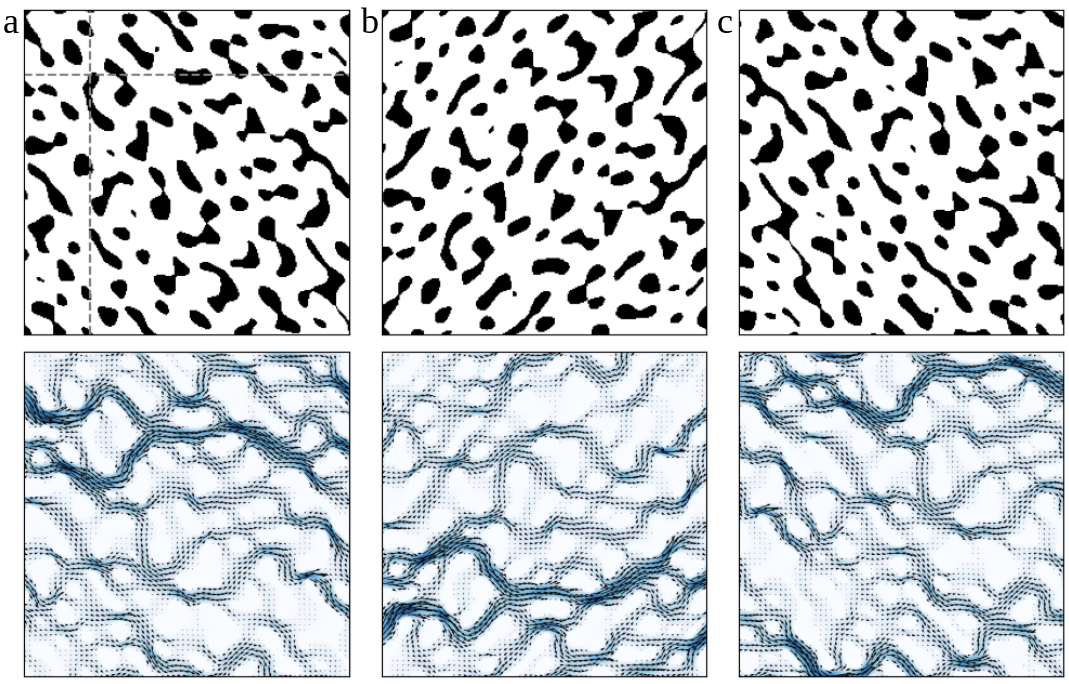}
    \caption{a) Example porous structure (top row) and corresponding LBM solution - color indicates velocity magnitude. b) The same structure and velocity field after vertical flip is applied. c) Effect of horizontal and vertical roll: 20\% of image rows and columns (indicated by dashed lines in panel a) where shifted beyond the original image and placed at the other end.}
    \label{fig:augmentations}
\end{figure}
During training, data augmentation techniques are applied to the structures to enhance the model's generalization ability. The augmentations include random vertical flipping and random rolling.  
In the vertical flip augmentation, each structure is flipped along the vertical axis with a given probability (set to $0.5$ in all experiments). The corresponding velocity field is adjusted accordingly by inverting the vertical velocity component ($v_y \rightarrow -v_y$), see Figure~\ref{fig:augmentations}b.
The random roll augmentation shifts a selected number of columns from the left side of the structure to the right and similarly moves a number of top rows to the bottom. The velocity field is adjusted consistently to maintain continuity. The effect is shown in Figure~\ref{fig:augmentations}c. 
The number of shifted columns and rows is randomly selected for each instance, ensuring that no more than 30\% of the image is rolled.  

These augmentations allow the model to learn from a more diverse set of training samples, effectively increasing the variability of the dataset without requiring additional computationally expensive LBM simulations.

\subsection*{Model architecture}
The solution employs a U-Net architecture \cite{unet_paper}, a widely adopted neural network originally developed for pixel-wise image segmentation. U-Net features an encoder-decoder structure with skip connections, which facilitate the preservation of spatial details while enabling the extraction of hierarchical, multi-scale features. 
\modified{
The input to the model is a $256\times256$ binary image representing the porous structure. The image is processed through the encoder path, which is composed of a deep CNN backbone. In our study, we consider a range of architectures commonly used in computer vision, including classical models such as VGG \cite{vgg_paper}, ResNet \cite{resnet_paper}, and DenseNet \cite{densenet_paper}, as well as more recent designs such as ConvNeXt \cite{liu2022convnet2020s}, EfficientNet \cite{tan2019efficientnet}, and MobileNetV3~\cite{howard2019searchingmobilenetv3}. These newer architectures have been proposed as either modernized or computationally efficient variants of standard convolutional networks. In particular, ConvNeXt incorporates several design elements inspired by transformer-based models while remaining fully convolutional, EfficientNet introduces a compound scaling strategy, and MobileNetV3 focuses on lightweight computation through the use of depthwise separable convolutions and optimized building blocks. All these architectures are used interchangeably as encoder backbones within the same framework, allowing for a consistent comparison.
}

Next, a bottleneck layer refines the internal feature representation before upsampling. The decoder consists of transposed convolutional layers that successively restore spatial resolution. Additionally, skip connections from corresponding encoder layers are incorporated, mitigating information loss due to downsampling. These skip connections enhance the network's ability to produce pixel-wise velocity predictions. 

The final output is a $256 \times 256$ velocity field with two channels, corresponding to the $x$ and $y$ components of the predicted velocity field ($u_x$ and $u_y$). This prediction is compared to the reference velocity field obtained via LBM simulation, and the discrepancy is quantified using the custom loss function defined in Equation~\eqref{eqn:loss_total}. The loss is then backpropagated through the network to update the model parameters.

\subsection*{Learning procedure}\label{seclearn}
For each porous structure in the dataset, the LBM solution was computed. The solution at each pixel is represented as a two-dimensional velocity vector $\mathbf{u} = (u_x, u_y)$. During the network training, the model receives the structure (binary image) as input and the corresponding LBM solution (pixel-size vector field) as output. The goal of training is to predict the LBM velocity field exclusively from the provided structure image of the structure. 

A notable feature of our approach is that the model backbone, which we use to construct the U-Net architecture, consists of convolutional neural networks (CNNs) of different architectures. Since CNNs operate as pattern recognition models, they are not explicitly aware of solving a differential equation. Instead, the network learns a mapping between the input structure and the output velocity field.  
To train the model, we employ a custom loss function $\mathcal{L}$, which is composed of multiple terms:  

\modified{
\begin{align}
\mathcal{L} &=  \mathcal{L}_\text{vel} 
+ \alpha \mathcal{L}_\text{obstacle} 
+ \beta \mathcal{L}_\text{div} 
+ \gamma \mathcal{L}_\text{perio}  
+ \delta \mathcal{L}_\text{tort}
\label{eqn:loss_total} \\
&= 
\frac{1}{2|\Omega|}
\sum_{j \in \{x, y\}} 
\int_{\Omega} 
\left( v_j(\xb) - u_j(\xb) \right)^2 d\xb  
+  
\frac{\alpha}{2|\mathcal{B}|}
\sum_{j \in \{x, y\}} 
\int_{\mathcal{B}} 
|u_j(\xb)| d\xb  \\
&+ 
\frac{\beta}{|\mathcal{P}|}
\int_{\mathcal{P}} 
(\nabla \cdot \mathbf{u}(\xb))^2 d\xb  
\\
&+ 
\frac{\gamma}{2|\Omega|}
\sum_{j \in \{x, y\}} 
\int_{\Omega} 
\left( u_j(\xb) - u_j^T(\xb) \right)^2 d\xb 
+ 
\delta(\tau_v-\tau_u)^2 .
\nonumber
\end{align}

where $\Omega$ denotes the full computational domain and $|\Omega|$ its area, $\mathcal{B}$ denotes the solid obstacle region and $|\mathcal{B}|$ its area, and $\mathcal{P}$ denotes the fluid pore space and $|\mathcal{P}|$ its area. Furthermore, $\mathbf{v}=(v_x, v_y)$ is the reference velocity field obtained from the LBM solution, $\mathbf{u}=(u_x, u_y)$ is the velocity field predicted by the model, and $\tau_v$ and $\tau_u$ are the reference and predicted tortuosities, respectively. The coefficients $\alpha$, $\beta$, $\gamma$, and $\delta$ are non-negative weights controlling the relative contributions of the individual loss components. All spatial integrals are evaluated numerically on the discrete $256\times256$ computational grid.
}

The loss function can be decomposed into five contributions. 
The first term, $\mathcal{L}_\text{vel}$, represents the mean squared error between the predicted and reference velocity fields. The error is computed separately for each velocity component and then summed.  
The second term, $\mathcal{L}_\text{obstacle}$, imposes an additional penalty when nonzero flow velocity is predicted inside the obstacle regions. Specifically, within the solid domain $\mathcal{B}$ (as defined in Equation~(\ref{eqn:random_threshold})), the penalty is applied using the L1 norm instead of the L2 norm employed in the velocity loss term. The use of the L1 norm places greater emphasis on reducing even small nonzero velocity predictions inside obstacles.  

The third term, $\mathcal{L}_\text{div}$, penalizes divergence in the predicted velocity field. In low-Mach number flows (which is our case), LBM solutions should satisfy the incompressibility condition $\nabla \cdot \mathbf{u} = 0$ everywhere. This term ensures that the predicted velocity field adheres to this constraint by minimizing divergence errors.  

\begin{figure}
    \centering
    \includegraphics[width=0.95\linewidth]{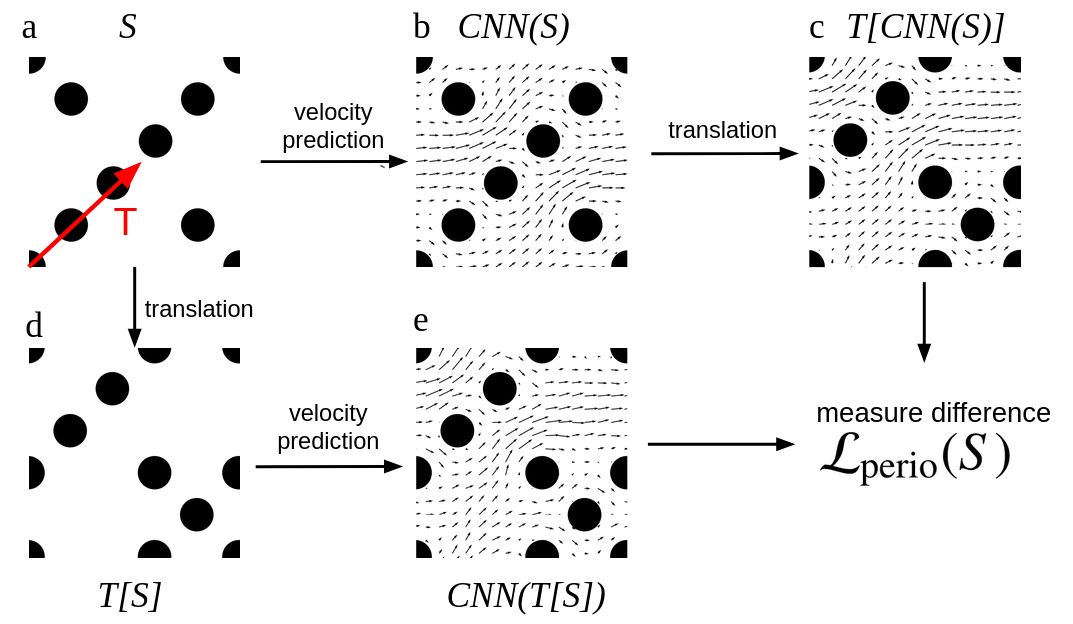}
    \caption{Explanation of the loss term $\mathcal{L}_\text{perio}$ defined in equation (\ref{eqn:loss_perio}). a) The structure $S$ and translation vector $T$. b) Velocity field predicted for structure $S$ -- this velocity field is denoted as $CNN(S)$. c) Velocity field after imposing translation $T$. d) Structure $S$ after translation by vector $T$: i.e. $T[S]$. e) Velocity field $CNN(T[S])$ predicted for translated structure $T[S]$. Finally $\mathcal{L}_\text{perio}$ is computed as integrated squared difference between velocity fields predicted for transformed structure $T[S]$ and transformed velocity field predicted for initial structure $S$.}
    \label{fig:period}
\end{figure}

The fourth term, $\mathcal{L}_\text{perio}$, addresses the periodicity of the system. 
Convolutional neural networks (CNNs), by design, are not inherently aware of periodic boundary conditions. They typically implement border treatment via padding methods—such as zero, replicate, reflect, or circular—which are selected heuristically and do not enforce any physical periodicity. Several studies demonstrate that the choice of padding can significantly influence prediction accuracy, and that even circular padding fails to generalize across periodic domains without additional architectural constraints or context information \cite{Innamorati2019, Gao2021, Alguacil2021}.
To mitigate this limitation and foster consistency under periodic shifts, we introduce an additional loss term $\mathcal{L}_\text{perio}$ that penalizes discrepancies between prediction for translated structure $T[S]$ and translated prediction for structure $S$. The mechanism by which periodic consistency is promoted through this loss term is illustrated in Figure~\ref{fig:period}. The translation $T$ takes into account periodic conditions. 
For a given structure $S$ and periodic translation $T$, this term is defined as
\modified{
\begin{equation}
\mathcal{L}_\text{perio}
=
\frac{1}{2|\Omega|}
\sum_{j \in \{x,y\}}
\int_{\Omega}
\left(
u_j(\xb) - u_j^T(\xb)
\right)^2
d\xb .
\label{eqn:loss_perio}
\end{equation}
Here, $\mathbf{u}^T$ denotes the prediction obtained after applying a periodic translation to the input structure and translating the resulting predicted velocity field back to the original coordinate system.
}
This encourages the network to produce predictions that are more consistent across periodic boundaries and better aligned with the periodic assumptions used in the LBM simulations. During training, several choices of the translation vector $T$ were tested, including random translations. In the final implementation, we used $T=[L/2,L/2]$, which provided a practical compromise between training cost and prediction accuracy.

Finally, the fifth term, $\mathcal{L}_\text{tort}$, penalizes discrepancies between the predicted tortuosity $\tau_u$ and the reference tortuosity $\tau_v$, which is computed from the ground-truth LBM solution. Tortuosity is a key macroscopic transport property that quantifies the effective elongation of flow paths due to the geometry of the porous medium. Accurate estimation of tortuosity is crucial in many real-world applications—including filtration, catalysis, and subsurface flow—where it directly impacts performance and efficiency metrics. By explicitly incorporating tortuosity into the loss function, we ensure that the model not only captures local velocity features but also respects global structural characteristics of the flow field, thereby enhancing the reliability of the model for downstream physical analysis and design tasks.

All the integrals in Equations (\ref{eqn:loss_total}) and (\ref{eqn:loss_perio}) were approximated by discrete sums over all pixels. Similarly the divergence in $\mathcal{L}_\text{div}$ i.e. $\nabla \cdot \mathbf{u} = \frac{\partial u_x}{\partial x} + \frac{\partial u_y}{\partial y}$ was approximated by first-order discrete derivatives. 

\modified{
We emphasize that the model is trained in a supervised manner using velocity fields generated by the LBM solver. Thus, the network should be interpreted as a surrogate model for the numerical solution under the considered assumptions, rather than as a solver that independently derives the governing equations. In this work, the term physics-informed refers to the use of additional physically motivated penalty terms in the loss function, which guide the supervised learning process toward predictions that better satisfy local and global physical constraints.
}

\subsection*{Components of the loss function - analysis}
\begin{table*}
\centering
\caption{Performance metrics for models trained with different combinations of loss function components. Each model shares the same ResNet-101 architecture and dataset. The table reports the average value of each individual loss component as well as the coefficient of determination ($R^2_\tau$) for tortuosity prediction}
\begin{tabular}{rrrr|rrrrrr}
\multicolumn{1}{l}{$\alpha$} & \multicolumn{1}{l}{$\beta$} & \multicolumn{1}{l}{$\gamma$} & \multicolumn{1}{l}{$\delta$} & \multicolumn{1}{l}{$\mathcal{L}_\text{vel}$} & \multicolumn{1}{l}{$\mathcal{L}_\text{obstacle}$} & \multicolumn{1}{l}{$\mathcal{L}_\text{div}$} & \multicolumn{1}{l}{$\mathcal{L}_\text{perio}$} & \multicolumn{1}{l}{$\mathcal{L}_\text{tort}$} & \multicolumn{1}{l}{$R^2_\tau$} \\ \hline
\multicolumn{1}{l}{0} & \multicolumn{1}{l}{0} & \multicolumn{1}{l}{0} & \multicolumn{1}{l}{0} & \multicolumn{1}{l}{2.26E-05} & \multicolumn{1}{l}{4.68E-04} & \multicolumn{1}{l}{6.68E-06} & \multicolumn{1}{l}{3.27E-05} & \multicolumn{1}{l}{6.22E-03} & \multicolumn{1}{l}{0.127} \\
\multicolumn{1}{l}{5} & \multicolumn{1}{l}{0} & \multicolumn{1}{l}{0} & \multicolumn{1}{l}{0} & \multicolumn{1}{l}{3.68E-05} & \multicolumn{1}{l}{4.43E-06} & \multicolumn{1}{l}{6.41E-06} & \multicolumn{1}{l}{3.34E-05} & \multicolumn{1}{l}{6.85E-04} & \multicolumn{1}{l}{0.901} \\
\multicolumn{1}{l}{5} & \multicolumn{1}{l}{1} & \multicolumn{1}{l}{0} & \multicolumn{1}{l}{0} & \multicolumn{1}{l}{2.00E-05} & \multicolumn{1}{l}{1.45E-06} & \multicolumn{1}{l}{2.00E-06} & \multicolumn{1}{l}{2.30E-05} & \multicolumn{1}{l}{1.09E-03} & \multicolumn{1}{l}{0.833} \\
\multicolumn{1}{l}{5} & \multicolumn{1}{l}{1} & \multicolumn{1}{l}{0.1} & \multicolumn{1}{l}{0} & \multicolumn{1}{l}{1.79E-05} & \multicolumn{1}{l}{1.42E-06} & \multicolumn{1}{l}{1.84E-06} & \multicolumn{1}{l}{1.86E-05} & \multicolumn{1}{l}{8.71E-04} & \multicolumn{1}{l}{0.873} \\
\multicolumn{1}{l}{5} & \multicolumn{1}{l}{1} & \multicolumn{1}{l}{0.1} & \multicolumn{1}{l}{0.01} & \multicolumn{1}{l}{2.07E-05} & \multicolumn{1}{l}{1.85E-06} & \multicolumn{1}{l}{2.09E-06} & \multicolumn{1}{l}{2.22E-05} & \multicolumn{1}{l}{1.24E-04} & \multicolumn{1}{l}{0.983} \\
\end{tabular}
\label{tab:loss_function_components}
\end{table*}

\begin{figure}
    \centering
    \includegraphics[width=0.99\linewidth]{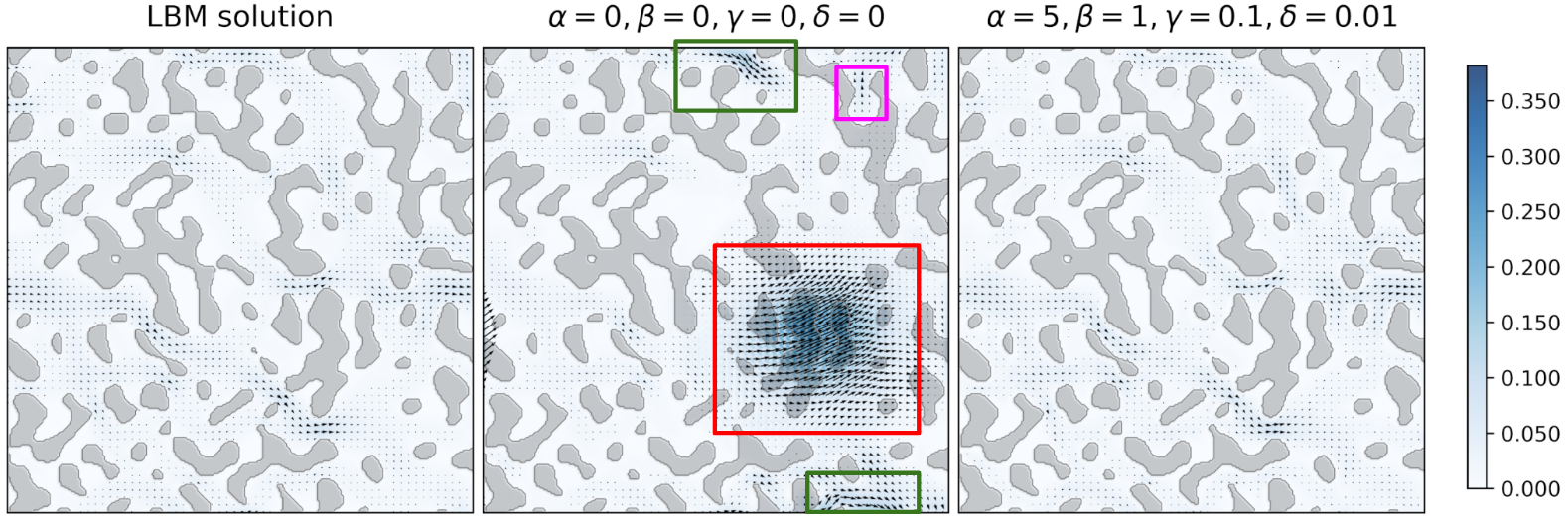}
    \caption{(left) Reference velocity field obtained from the LBM simulation. (middle) Prediction from a model trained with a minimal loss function configuration ($\alpha=\beta=\gamma=\delta=0$), illustrating several failure modes. The red box indicates flow erroneously penetrating solid obstacles. The green box highlights spurious flow induced by the model’s failure to respect periodic boundary conditions. The purple box shows a region confined by obstacles on three sides, where the model incorrectly predicts the location of the outflow. These prediction artifacts are fixed when the full loss function is used, with optimized weights: $\alpha=5$, $\beta=1$, $\gamma=0.1$, and $\delta=0.01$ as shown in the right image.}
    \label{fig:loss_function_components}
\end{figure}

Table~\ref{tab:loss_function_components} presents a quantitative comparison of models trained with different configurations of the custom loss function. All models share the same ResNet-101 architecture, which was identified as the most performant among those evaluated, and were trained, validated, and tested on identical datasets. The values reported represent the average magnitude of each loss component on the test set. Additionally, the accuracy of tortuosity prediction is evaluated using the coefficient of determination $R^2_\tau$.
The baseline model was trained using only the velocity loss component, i.e., $\mathcal{L} = \mathcal{L}_\text{vel}$. This model achieved the lowest mean squared error (MSE) in terms of velocity prediction but failed to suppress spurious flow within obstacle regions, as illustrated in Figure~\ref{fig:loss_function_components}. Furthermore, it demonstrated poor accuracy in predicting tortuosity, with $R^2_\tau = 0.127$.

In the second experiment, we introduced the obstacle loss term $\mathcal{L}_\text{obstacle}$ with a weighting factor of $\alpha = 5$. This addition significantly reduced flow predictions within solid regions, with the average $\mathcal{L}_\text{obstacle}$ decreasing by over two orders of magnitude. It also led to a substantial improvement in tortuosity prediction, increasing $R^2_\tau$ to 0.901 and reducing the MSE in $\mathcal{L}_\text{tort}$ by one order of magnitude. However, the divergence loss  $\mathcal{L}_\text{div}$ remained similar due to the unconstrained divergence of the predicted field.
To address this, the third model incorporated the divergence loss term $\mathcal{L}_\text{div}$ with $\beta = 1$. This further reduced the average divergence in the predicted velocity fields and yielded improvements in both $\mathcal{L}_\text{vel}$ and $\mathcal{L}_\text{obstacle}$, indicating better generalization in void regions and at fluid–solid interfaces.
Despite these improvements, qualitative analysis (see Figure~\ref{fig:loss_function_components}) revealed that the model occasionally failed to enforce periodic boundary conditions. To address this issue, the fourth model introduced the periodic consistency loss $\mathcal{L}_\text{perio}$, weighted by $\gamma = 0.1$. This modification improved the model's ability to handle periodic translations, reducing $\mathcal{L}_\text{perio}$ and ensuring greater physical consistency across domain boundaries.
Finally, the fifth and most complete model incorporated the tortuosity consistency loss $\mathcal{L}_\text{tort}$, with a weighting factor of $\delta = 0.01$. This led to the most accurate tortuosity predictions, achieving an $R^2_{\tau}$ value of 0.983. Additionally, it exhibited the lowest penalties for both obstacle violation and divergence, while maintaining a velocity prediction error comparable to the baseline. The final model thus represents the best trade-off among local accuracy, physical plausibility, and global flow characteristics.

The above analysis illustrates the contribution of each individual loss function component to the overall model performance.
The sequential inclusion of these terms not only often improves local accuracy in velocity prediction but also enforces key physical constraints such as impermeability of solid regions, incompressibility, and periodic boundary conditions, while enhancing the fidelity of global transport metrics like tortuosity. Based on these findings, all subsequent models presented in this study were trained using the full loss function with fixed weighting coefficients: $\alpha = 5$, $\beta = 1$, $\gamma = 0.1$, and $\delta = 0.01$.

\modified{
A more detailed sensitivity analysis of these hyperparameters, including their interactions, is provided in the Supplementary Information (Section S3).
}

\subsection*{Training protocol}
During training, the dataset was randomly split into training, validation, and test sets, comprising 70\%, 15\%, and 15\% of the structures, respectively. The model's weights were initialized randomly.
We employed a two-stage training strategy. In the first stage, a pilot model was trained with divergence, periodicity and tortuosity contributions to total loss set to zero, i.e. $\beta=\gamma=\delta=0$ in (\ref{eqn:loss_total}). In the second stage, the pilot model's weights were used as a starting point and were then fine-tuned using the full loss formulation with final values of hyperparameters, i.e. $\beta=1, \gamma=0.1$ and $\delta=0.01$. This staged approach improved training stability and reduced sensitivity to the random dataset split.
To prevent data leakage, both the pilot and final models were trained using the exact same data split. Optimization was carried out using the Adam algorithm \cite{kingma2017}. For the pilot model, a learning rate of 0.0005 and gradient clipping \cite{pascanu2013} with a threshold of 1.0 were used. For the final model, the learning rate was reduced to $10^{-5}$ with gradient clipping at 0.1. In both cases, training continued as long as loss measure on validation set decreases (early stopping technique). 
\modified{Further training details are provided in the Supplementary Information (Section S2).}

Only test set metrics are reported in the paper. To improve robustness, three different random data splits were evaluated for each model architecture, and the reported results in Table \ref{tab:results} reflect the average and standard deviation across these splits.

\section*{Results}\label{secRES}
\begin{table*}[]
\centering
\caption{
Performance metrics for various evaluated neural network architectures. The pixel-wise root mean squared error (RMSE) is reported for velocity field predictions. For tortuosity and permeability, three metrics are provided: RMSE, mean absolute percentage error (MAPE), and coefficient of determination ($R^2$). The best result for each metric is highlighted in \textbf{bold}, second-best with \textbf{\textit{bold+italic}}, while the thrid best is shown in \textit{italic}. For velocity RMSE standard deviation is reported based on three independent runs with different dataset splits.}
\modified{
\begin{tabular}{l|rr|rrr|rrr}
 & \multicolumn{2}{c|}{velocity} & \multicolumn{3}{c|}{tortuosity}& \multicolumn{2}{c}{permeability} &\\
 & {RMSE} & {std(RMSE)} & {RMSE} & {MAPE} & {$R^2$} & {RMSE} & {MAPE} & {$R^2$} \\ \hline
VGG16 & 1.09E-02 & \textbf{5.62E-04} & 1.91E-02 & 1.0 & 0.949 & 1.87E+03 & 9.1 & 0.985 \\
ResNet18 & 1.07E-02 & 2.20E-03 & 2.70E-02 & 1.5 & 0.897 & 2.67E+03 & 17.2 & 0.971 \\
ResNet50 & 8.58E-03 & 2.89E-03 & \textit{1.89E-02} & 1.1 & 0.950 & \textit{1.53E+03} & \textit{7.0} & 0.990 \\
ResNet101 & \textbf{5.10E-03} & \textbf{\textit{1.92E-03}} & \textbf{1.13E-02} & \textbf{0.6} & \textbf{0.983} & \textbf{8.69E+02} & \textbf{4.8} & \textbf{0.997} \\
ResNet152 & 7.74E-03 & 2.95E-03 & 2.01E-02 & 1.0 & 0.944 & 1.89E+03 & 11.8 & 0.985 \\
DenseNet121 & \textbf{\textit{6.45E-03}} & 5.71E-03 & 4.11E-02 & 2.9 & 0.763 & 1.70E+03 & 10.5 & 0.988 \\
DenseNet201 & \textit{7.48E-03} & 2.93E-03 & \textbf{\textit{1.81E-02}} & \textit{0.9} & \textit{0.954} & 2.03E+03 & 13.2 & 0.983 \\
ConvNext Tiny & 7.74E-03 & 2.45E-03 & 2.07E-02 & 1.1 & 0.940 & 1.55E+03 & 8.6 & 0.990 \\
ConvNext Small & 9.24E-03 & \textit{1.65E-03} & 2.41E-02 & 1.3 & 0.919 & 1.71E+03 & 9.9 & 0.987 \\
ConvNext Base & 4.40E-02 & 1.94E-02 & 2.22E-02 & 1.2 & 0.928 & \textbf{\textit{1.47E+03}} & 8.9 & \textit{0.991} \\
EfficientNet B2 & 1.20E-02 & 3.18E-03 & 2.40E-02 & 1.4 & 0.919 & 1.93E+03 & 13.9 & 0.983 \\
EfficientNet V2 S & 1.15E-02 & 4.51E-03 & 2.86E-02 & 1.6 & 0.886 & 2.41E+03 & 12.8 & 0.975 \\
MobileNet V3 Small & 1.00E-02 & 1.90E-03 & 2.49E-02 & 1.3 & 0.913 & 1.81E+03 & 10.6 & 0.985 \\
MobileNet V3 Large & 6.60E-03 & 3.57E-03 & \textit{1.68E-02} & \textbf{\textit{0.9}} & \textbf{\textit{0.961}} & \textit{1.26E+03} & \textbf{\textit{7.1}} & \textbf{\textit{0.993}} \\
\end{tabular}
}
\label{tab:results}
\end{table*}

Table~\ref{tab:results} presents the predictive performance metrics for all evaluated architectures, \modified{covering a range of convolutional backbones with varying levels of architectural complexity and computational efficiency.} The accuracy of the velocity field predictions is assessed using the component-wise, pixel-averaged root mean squared error (RMSE) defined as $\textrm{RMSE}=\sqrt{\textrm{MSE}}$. To provide additional context for these values, Figure~\ref{fig:summaries}b shows the distributions of both the mean and maximum fluid velocities in the dataset. The std(RMSE) column show the standard deviation of that RMSE computed out of three independent runs with different dataset splits. 
The next columns show the performance metrics, i.e. the RMSE, Mean Absolute Percentage Error (MAPE) and coefficient of determination $R^2$ for tortuosity and permeability computed from the velocity field predicted by the model and compared with the LBM solution. 

Among all tested models, ResNet-101 consistently achieves the highest performance across nearly all metrics. It yields the lowest RMSE for velocity predictions at $5.1 \cdot 10^{-3}$, which corresponds to the average magnitude of the prediction error per velocity component. Given the median fluid velocity of approximately $2.5 \cdot 10^{-2}$ (Figure~\ref{fig:summaries}b), this error reflects high predictive accuracy. Moreover, the low std(RMSE) value for ResNet-101 indicates robustness and stable generalization across dataset variations.

\modified{Architectures designed with a focus on computational efficiency, such as EfficientNet and MobileNetV3, tend to exhibit higher errors across both velocity reconstruction and derived physical quantities. ConvNeXt, which follows a more modern convolutional design, achieves performance comparable to some of the classical architectures, but does not surpass the best-performing models. This trend is consistent across the evaluated metrics, including RMSE, MAPE, and $R^2$.}

\begin{figure}
    \centering
    \includegraphics[width=\linewidth]{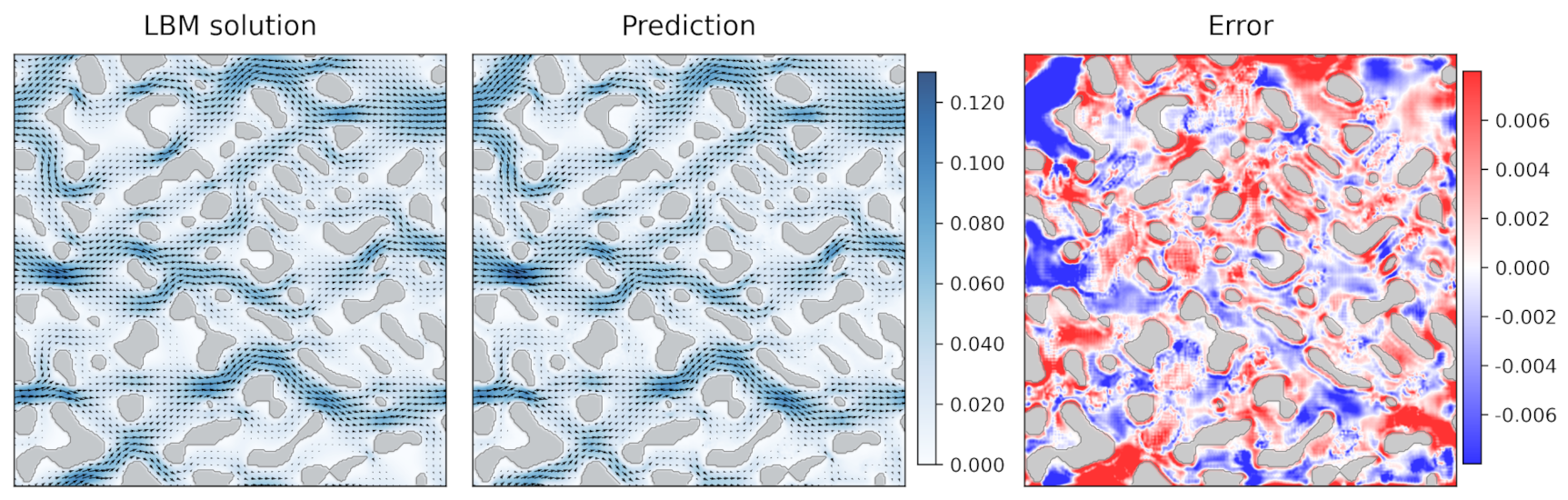}
    \caption{(left) Reference velocity field obtained from LBM simulation. (middle) Prediction from model. (right) Difference in velocity magnitude between reference and predicted velocity fields. Results for a selected, representative sample for which the RMSE prediction 
of the velocity field is 5.8E-03, minimally greater than the average RMSE of the entire test-set of the model (5.1E-03, c.f. Table \ref{tab:results}).}
    \label{fig:lbm_cnn_error}
    
\end{figure}

In Fig.~\ref{fig:lbm_cnn_error}, we present the prediction results for a selected porous sample in graphical form. In addition to the velocity field from the LBM method and from the ResNet-101 prediction, we also show the difference in the 
absolute value of the velocity field. A visual analysis of these results confirms that the CNN network can 
correctly predict fluid flow paths. Analysis of the error distribution map did not yield any systematic conclusions. 
The differences between two methods are not significant and fluctuate around a value of 0, not 
betraying any trend towards one of the methods. It is worth noting the excellent continuity of the system at the periodic edges, which is a result of the periodic term in the loss function introduced by us.
\modified{Additional error-bound analysis is provided in the Supplementary Information (Section S1).}

\begin{figure}
    \centering
    a \includegraphics[width=0.45\linewidth]{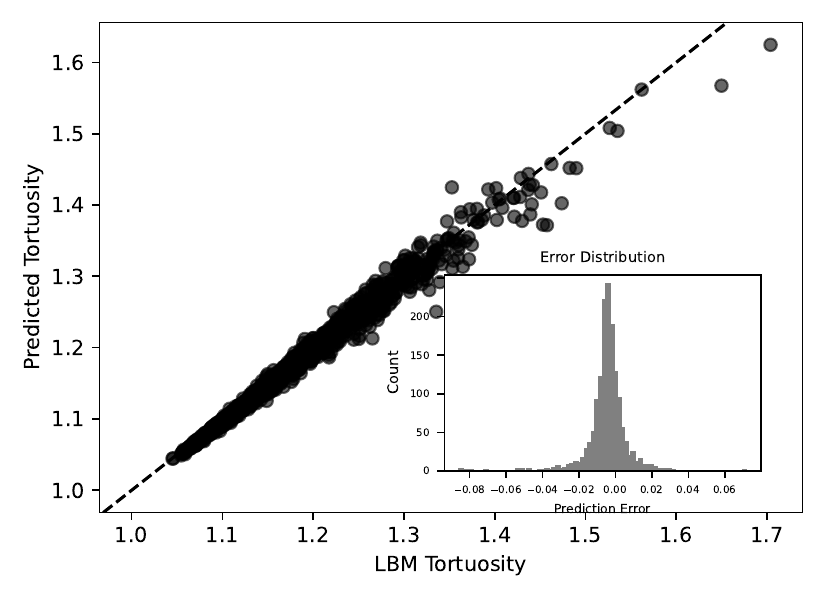}
    b \includegraphics[width=0.45\linewidth]{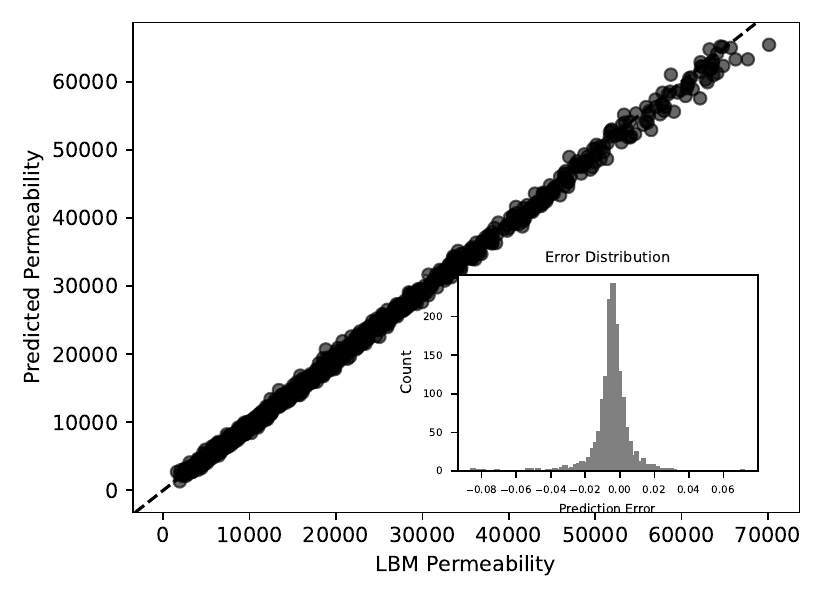}
    \caption{a) Tortuosity predicted by the best-performing model as the function of LBM computed tortuosity. Inset shows error distribution. b) The same for permeability.}
    \label{fig:tort_perm_errorplot}
\end{figure}

Based on the velocity fields, the tortuosity and permeability of each structure were inferred and compared with the LBM ground truth. For tortuosity, ResNet-101 achieves an RMSE of $1.13 \cdot 10^{-2}$—small relative to the mean dataset value of 1.1719 (Figure~\ref{fig:summaries}d)—and a remarkably low MAPE of 0.6\%. This indicates that, on average, the predicted tortuosity deviates by less than 1\% from the reference. As shown in Figure~\ref{fig:tort_perm_errorplot}a, the predicted tortuosity values align closely with the exact LBM values, lying along the diagonal with only minor deviations at higher tortuosity values. The associated error distribution is symmetric and tightly centered around zero.

For permeability, ResNet-101 again demonstrates the best performance, achieving the lowest RMSE and MAPE, and the highest $R^2$ across all models. The MAPE remains below 5\%, and the RMSE of $8.69 \cdot 10^2$, when compared to the average dataset permeability of $2.3 \cdot 10^5$ (Figure~\ref{fig:summaries}d), confirms the accuracy of the predictions. Figure~\ref{fig:tort_perm_errorplot}b further supports this, showing predicted permeability values tightly clustered around the true LBM values.

Overall, the results in Table~\ref{tab:results} demonstrate that, with the exception of ResNet-18, most architectures yield comparable performance. ResNet-50 consistently performs well and emerges as the second-best choice, balancing both velocity field prediction and macroscopic property estimation. The performance achieved by the DenseNet family of convolutional neural networks seems to be systematically lower than that achieved by the simpler ResNet architecture. Notably, the VGG16 architecture, despite its simpler design and lack of residual or dense connections, achieves competitive results—typically falling between the performance of the ResNet and DenseNet families. \modified{These results suggest that, for the task of mapping pore geometry to velocity fields on a fixed grid, increased architectural complexity or efficiency-oriented design does not necessarily translate into improved predictive accuracy.}

\modified{
A comparison with a Fourier Neural Operator~\cite{li2021fourier, kovachki2023neural} model is provided in the Supplementary Information (Section S4).
}

\subsection*{Generalization to different systems}
\begin{table*}[]
\centering
\caption{Performance metrics of the model on various datasets that are different from the one used to train the model.}
\begin{tabular}{l|r|rrr|rrr}
 & \multicolumn{1}{l|}{velocity} & \multicolumn{3}{c|}{tortuosity} & \multicolumn{3}{c}{permeability}  \\
 & RMSE & RMSE & MAPE & $R^2$ & RMSE & MAPE & $R^2$ \\ \hline
shape, $P_\text{circle}=1.0$ & 9.20E-03 & 1.2E-02 & 0.82 & 0.958 & 2.92E+03 & 4.39 & 0.994 \\
shape, $P_\text{circle}=0.5$ & 1.90E-02 & 1.7E-02 & 1.15 & 0.942 & 8.56E+03 & 4.59 & 0.981 \\
shape, $P_\text{circle}=0.0$ & 2.25E-02 & 2.2E-02 & 1.56 & 0.897 & 1.15E+04 & 5.60 & 0.976 \\ \hline
pipe & 5.59E-03 & 1.6E-02 & 0.76 & 0.963 & 9.18E+02 & 4.09 & 0.997 \\
pipe (central) & 6.39E-03 & 1.6E-02 & 0.79 & 0.962 & 8.49E+02 & 4.45 & 0.998 \\ \hline
porosity 0.7-0.8 & 3.52E-03 & 1.5E-02 & 0.76 & 0.958 & 5.70E+02 & 6.70 & 0.961 \\
porosity 0.6-0.7 & 2.64E-03 & 6.7E-02 & 2.50 & 0.794 & 3.36E+02 & 13.74 & 0.916 \\
porosity 0.5-0.6 & 1.86E-03 & 6.6E-01 & 12.32 & 0.116 & 2.65E+02 & 77.18 & 0.635 \\ \hline
\modified{Li-O$_2$ Electrodes} & 6.41E-03 & 3.4E-02 & 2.08 & 0.798 & 1.75E+03 & 9.71 & 0.986 
\end{tabular}
\label{tab:generalization}
\end{table*}

While the model is trained on a specific dataset of porous structures, generated using random trigonometric polynomials, the underlying architecture and training methodology can be applied to similar physical systems. The use of convolutional neural networks allows the model to learn spatially invariant features, enabling it to generalize to geometries and flow patterns that differ from those seen during training. Moreover, the inclusion of physical constraints in the loss function—such as incompressibility, zero-flow conditions inside solid regions, and periodicity—encourages the model to learn representations that are physically meaningful rather than dataset-specific. 
It is important to emphasize that all results presented in the following sections were obtained using the best-performing model, ResNet-101, trained solely on the original dataset. No further training, fine-tuning, or modification was performed for the out-of-distribution samples tested here, and the model had no prior exposure to these novel porous structures.
An important question we investigate in the following sections is whether this approach can extend to porous systems with substantially different porosity distributions, obstacle geometries, or boundary conditions, while maintaining predictive accuracy and physical plausibility. 
To this end, additional porous structures—distinct from those in the original training dataset—were generated to serve as out-of-distribution test cases. The velocity fields predicted by the model for these new samples were then compared with the corresponding LBM solutions to assess the model's generalization capabilities.
\subsubsection*{Obstacles of different shape}
\begin{figure}
    \centering
    \includegraphics[width=0.9\linewidth]{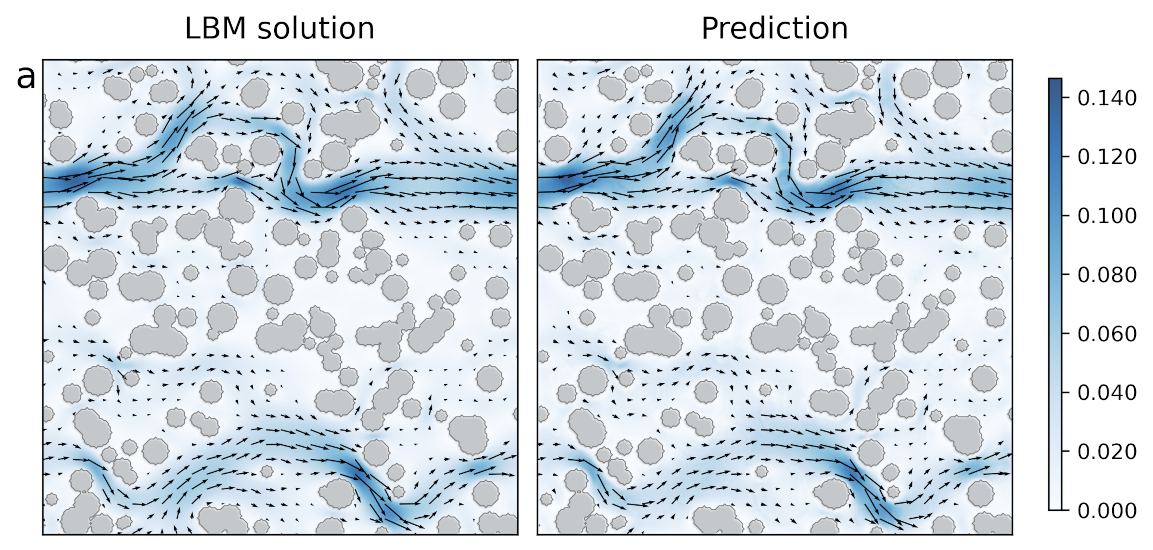}\\
    \includegraphics[width=0.9\linewidth]{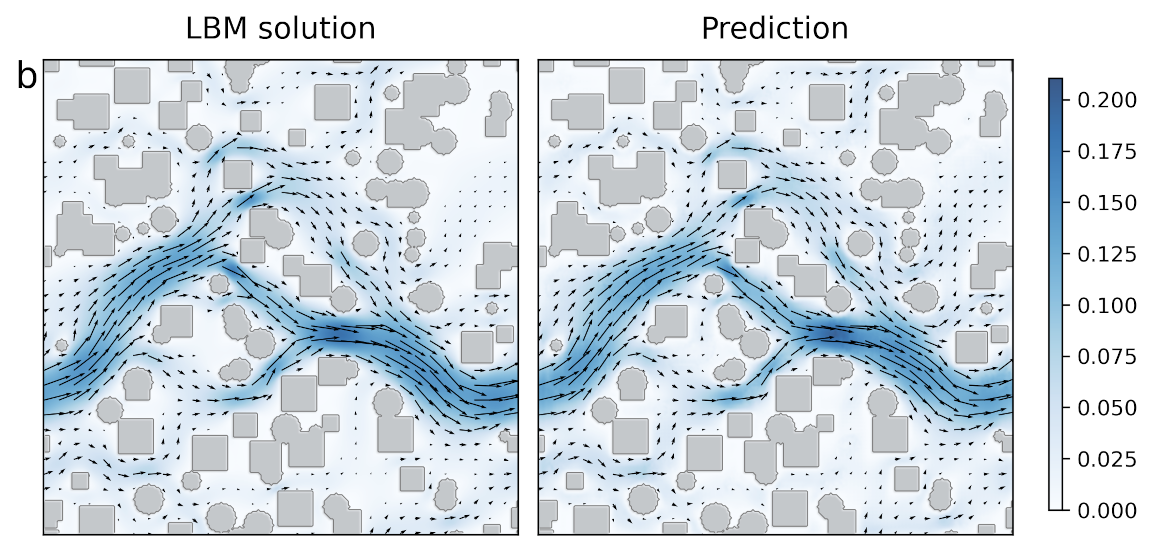}\\
    \includegraphics[width=0.9\linewidth]{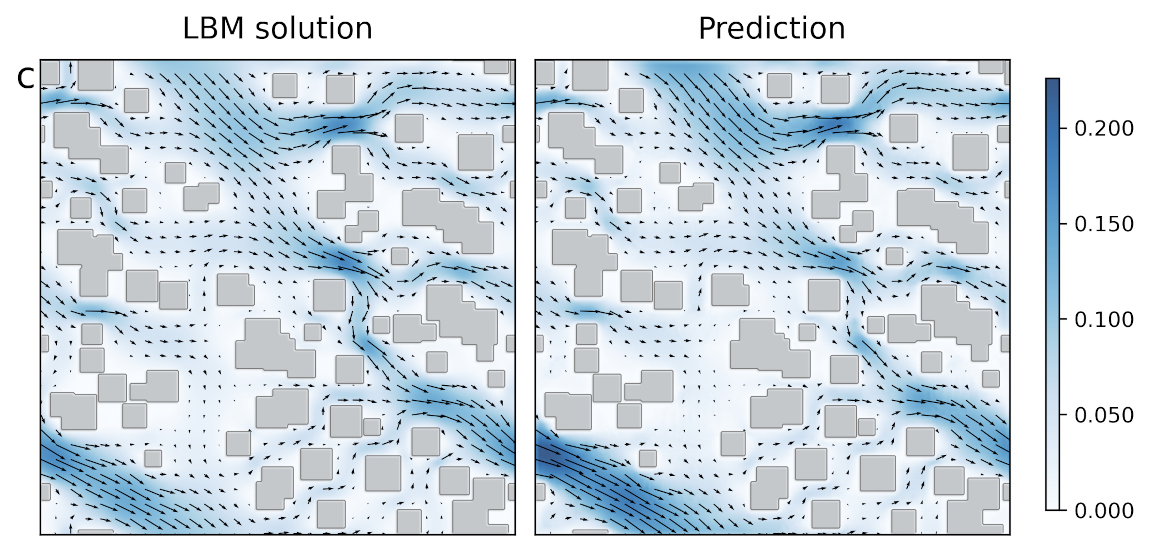}
    \caption{LBM computed and neural-network predicted velocity fields for exemplary systems with circular and square obstacles. a) Circle-only structures ($P_\text{circle}=1.0$), b) Mixed structures ($P_\text{circle}=0.5$), c) Square-only structures ($P_\text{circle}=0.0$). 
    }
    \label{fig:flow_shapes}
\end{figure}
A set of porous structures was generated using randomly placed circles and squares. Obstacle sizes were sampled uniformly between 3 and 8 pixels, and objects were added until the target porosity—randomly chosen between 0.70 and 0.95—was achieved. Each obstacle was a circle with probability $P_\text{circle}$ or a square with probability $1 - P_\text{circle}$. Three datasets were created with $P_\text{circle} = {0.00, 0.50, 1.00}$, corresponding to square-only, mixed, and circle-only configurations.
Figure~\ref{fig:flow_shapes} illustrates representative examples of the porous geometries, along with the corresponding LBM solutions and model predictions. Notably, despite the presence of obstacle shapes that differ from those used during model training, the predictions remain closely aligned with the LBM reference solutions, indicating encouraging generalization to this class of modified geometries.
As summarized in Table~\ref{tab:generalization}, the prediction quality shows a clear dependence on obstacle shape. Specifically, as the proportion of square obstacles increases, the accuracy of the predicted flow fields deteriorates. For the circle-only dataset ($P_\text{circle} = 1.00$), the model achieves performance comparable to that on the original training set, with mean squared errors (MSEs) of the same order of magnitude and high coefficients of determination: $R^2 = 0.958$ for tortuosity and $R^2 = 0.994$ for permeability. As the proportion of square obstacles increases, all performance metrics gradually degrade; however, even in the most challenging square-only case ($P_\text{circle} = 0.00$), the model maintains a mean absolute percentage error (MAPE) below 2\% for tortuosity and below 6\% for permeability, indicating overall robust generalization across varying geometric configurations.

\subsubsection*{Another boundary conditions}
\begin{figure}
    \centering
    \includegraphics[width=0.9\linewidth]{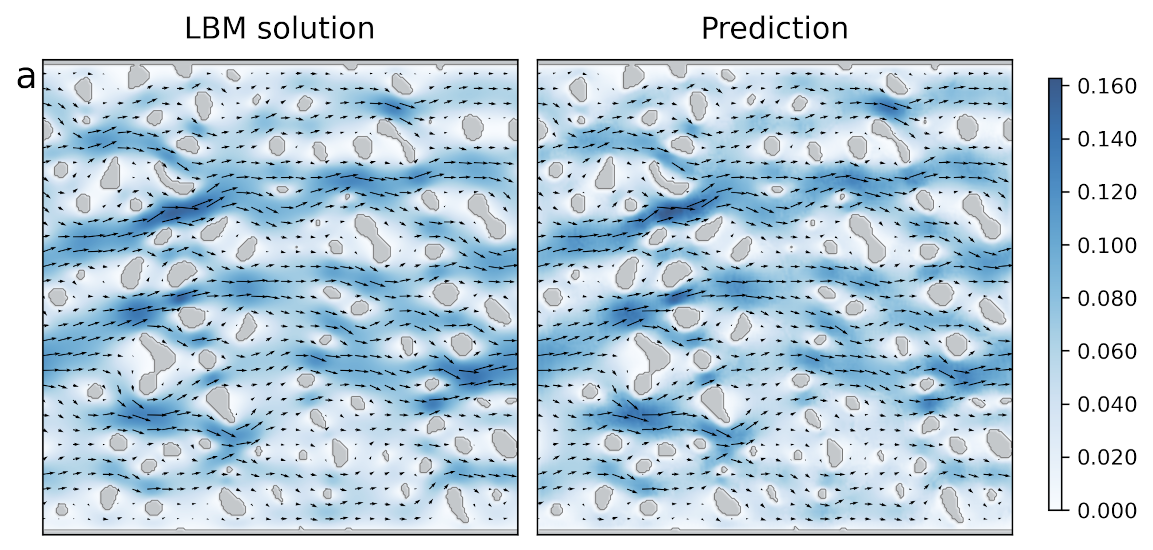}\\
    \includegraphics[width=0.9\linewidth]{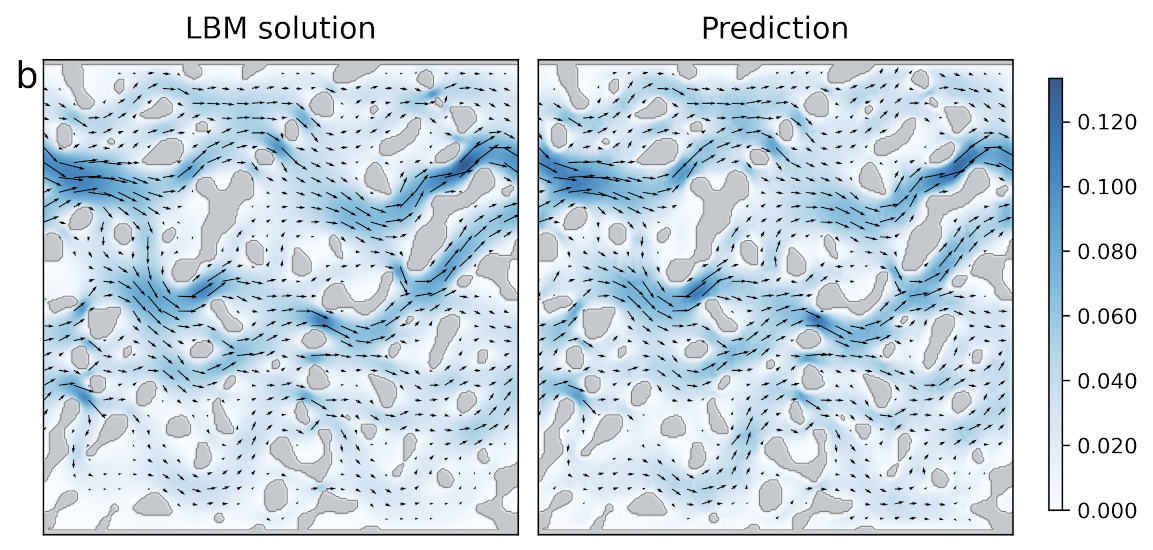}
    \caption{LBM computed and neural-network predicted velocities fields for exemplary porous systems constrained by two linear obstacles (pipe) at porosities a) 0.896 and b) 0.838.}
    \label{fig:flow_bonduaries_pipe}
\end{figure}

\begin{figure}
    \centering
    \includegraphics[width=0.9\linewidth]{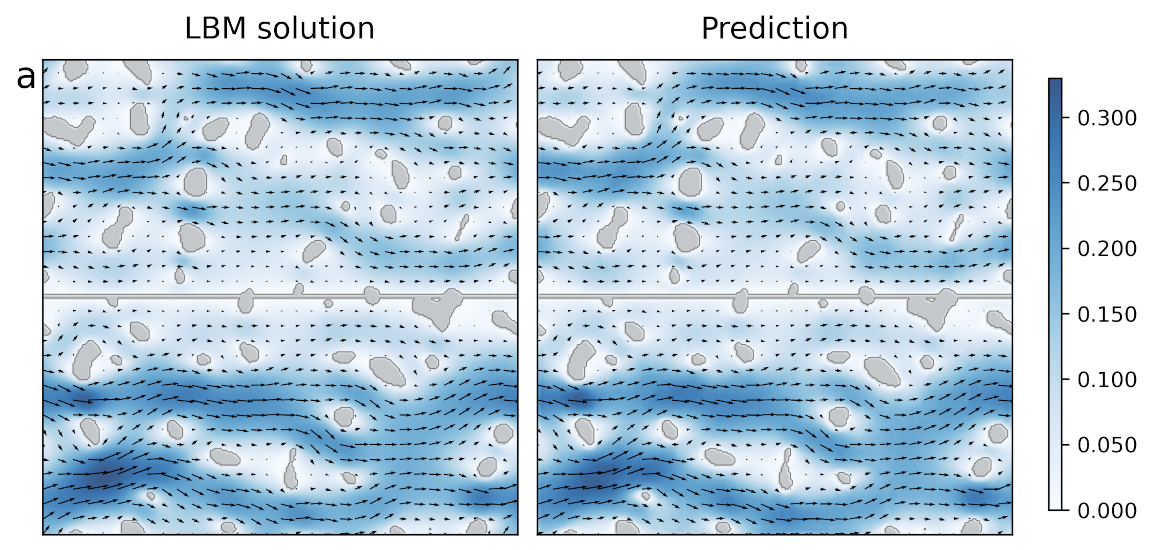}\\
    \includegraphics[width=0.9\linewidth]{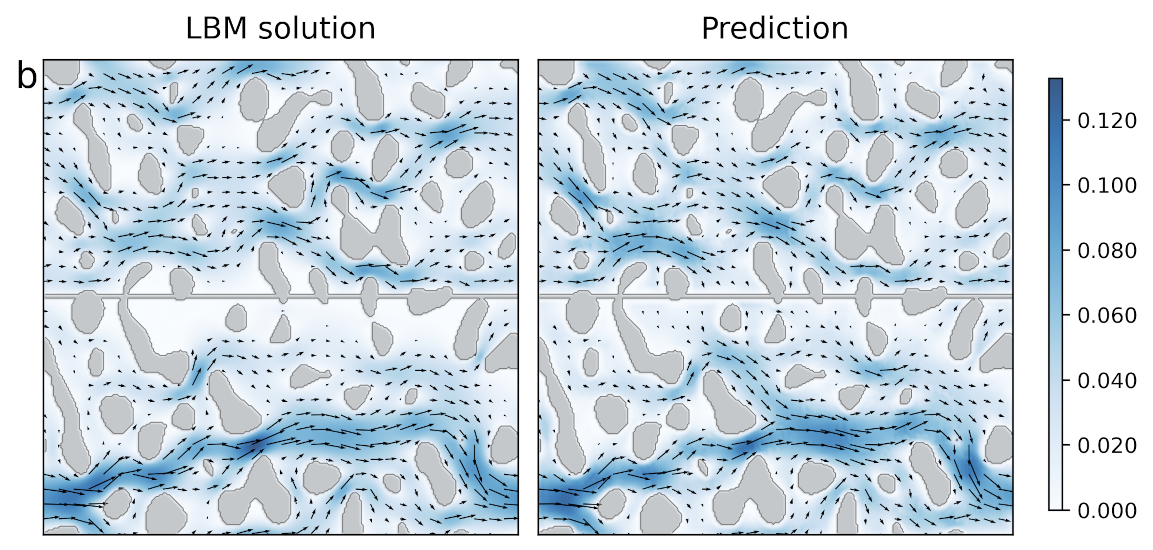}\\
    \caption{The same as Figure \ref{fig:flow_bonduaries_pipe} but with linear obstacle placed in the center of the system at porosities a) 0.927, b) 0.792.}
    \label{fig:flow_bonduaries_pipe_central}
\end{figure}

Periodic boundary conditions were modified by introducing solid vertical obstacles along the horizontal edges of the porous domain, aligned with the direction of fluid flow. While periodicity is preserved along the horizontal axis, the added boundaries effectively transform the system into an infinite channel or pipe reducing the effect of finite-size anisotropy and an inclination angle between external body force and resulting streamwise velocity \cite{Koza09}. Importantly, the model was not trained on systems with confined pipe-like geometries or with any constraint that explicitly prohibited flow across the horizontal edges.
Figure~\ref{fig:flow_bonduaries_pipe} shows the model’s predictions for three example systems with pipe-like geometries and periodic boundary conditions. Although the model was not explicitly trained on such configurations, it correctly reconstructs the overall flow profile.  Importantly, the velocity component perpendicular to the pipe axis remains negligible, demonstrating that the model implicitly respects the physical constraint of unidirectional flow in these settings. 

An equivalent physical setup can be achieved by placing a single solid barrier horizontally across the center of the sample. Although this configuration is physically analogous to side-wall confinement, it may pose a greater challenge to the model, as it relies more strongly on accurately capturing internal boundary effects rather than edge constraints alone -- some exemplary predictions made by the model are shown in Figure \ref{fig:flow_bonduaries_pipe_central}. 

As shown by the performance metrics in Table~\ref{tab:generalization}, the model maintains high predictive accuracy for the in-pipe systems, with only a slight reduction in performance compared to the original dataset. The coefficient of determination ($R^2$) exceeds 0.96 for tortuosity and 0.99 for permeability, while the mean absolute percentage error (MAPE) remains below 0.8\% and 4.5\%, respectively. These results indicate that the model generalizes well to previously unseen pipe-like geometries and \modified{continues to provide accurate predictions for both macroscopic transport properties in these test cases}.

\subsection*{Generalization to different porosities}
\begin{figure}
    \centering
    \includegraphics[width=0.9\linewidth]{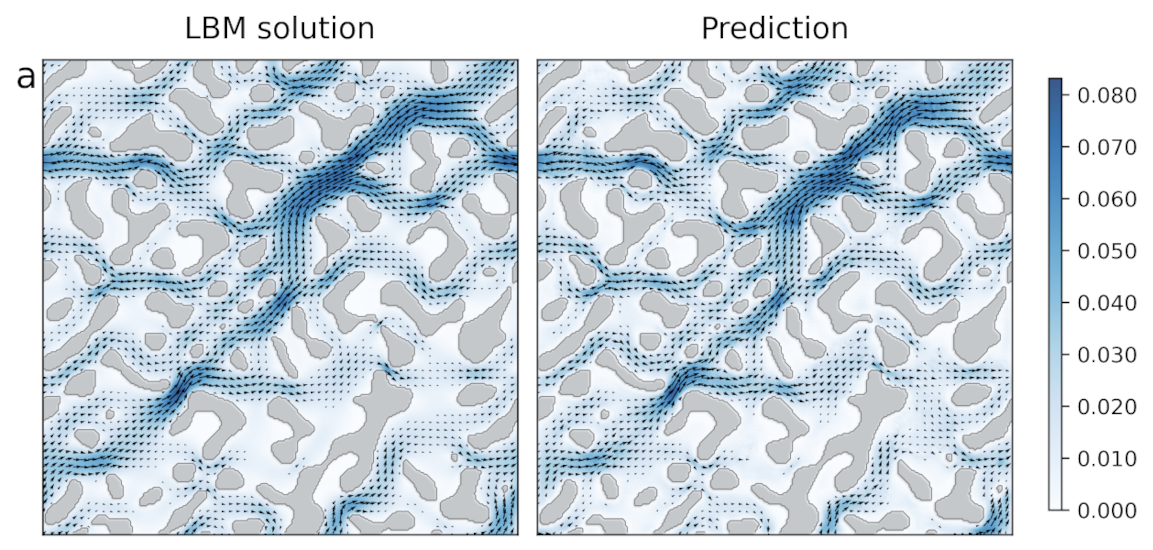}\\
    \includegraphics[width=0.9\linewidth]{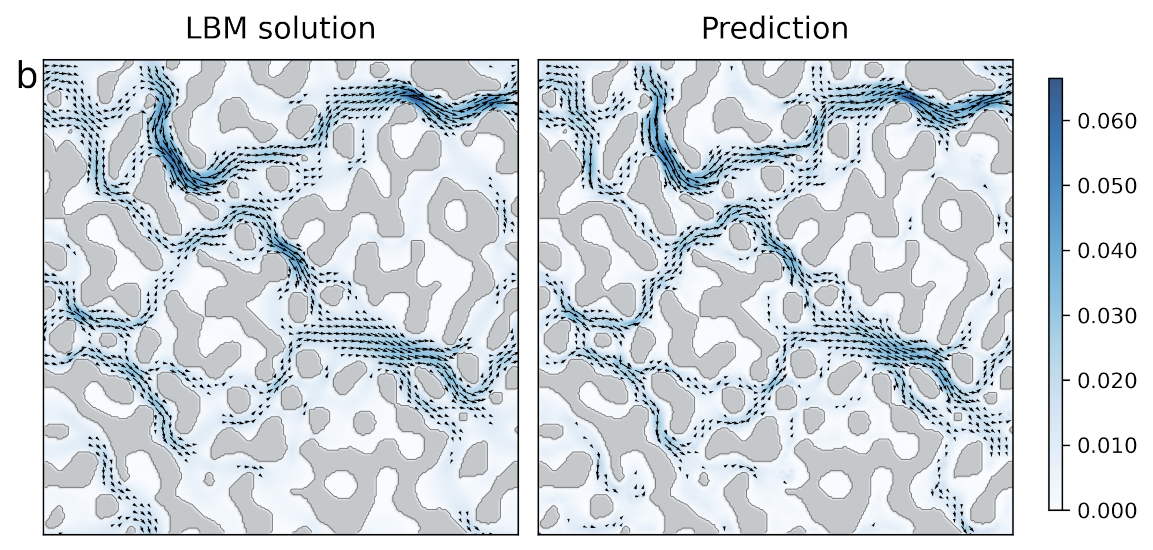}\\
    \includegraphics[width=0.9\linewidth]{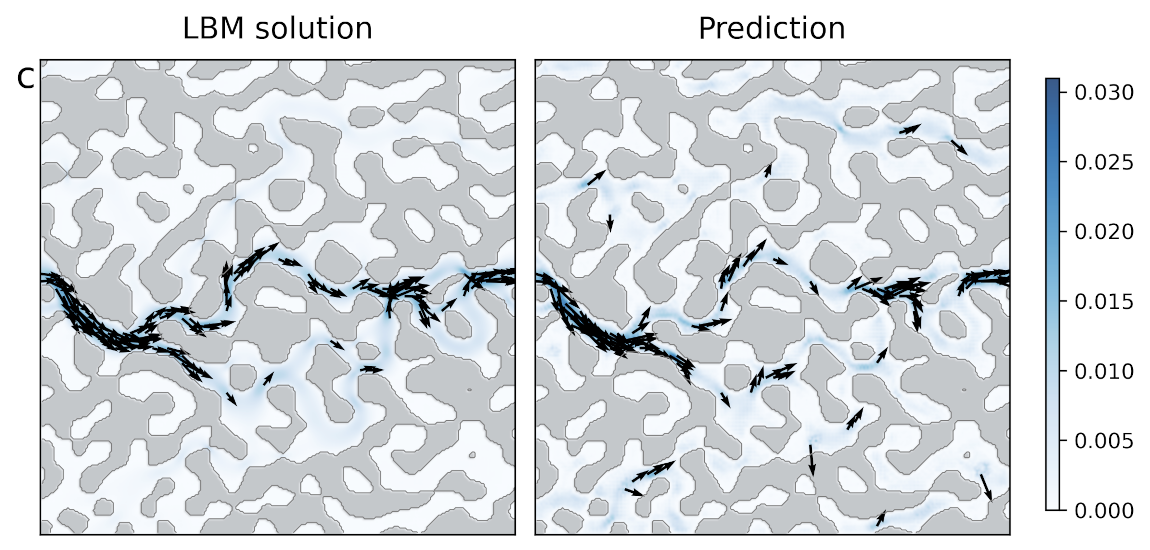}
    \caption{LBM computed and neural-network predicted velocities fields for exemplary porous systems of lower porosity. Systems have porosities: a) 0.752, b) 0.649, c) 0.512. }
    \label{fig:flow_porosity}
\end{figure}

Three additional datasets were constructed to evaluate the model’s performance on structures with decreasing porosity. The first dataset includes samples with porosity in the range $0.7 < \varphi < 0.8$, the second with $0.6 < \varphi < 0.7$, and the third with $0.5 < \varphi < 0.6$. Each dataset consists of 500 structures generated using the method defined in Equation~(\ref{eqn:random_trig}), with the threshold parameter $\varepsilon$ in Equation~(\ref{eqn:random_threshold}) adjusted to achieve the desired porosity. Figure~\ref{fig:flow_porosity} shows representative LBM solutions and corresponding predictions made by the model, each taken from a randomly selected sample within the respective datasets.

Performance metrics for each case are summarized in Table~\ref{tab:generalization}. The results reveal a clear dependency of model accuracy on porosity. As porosity decreases, prediction performance deteriorates—as reflected by declining $R^2$ scores and increasing MAPE for both tortuosity and permeability. Notably, the first dataset ($0.7\!<\!\varphi\!<\!0.8$) falls within the porosity range used during model training, and the model performs well on these samples. For the second dataset ($0.6\!<\!\varphi\!<\!0.7$), which lies outside the training distribution, the model still achieves reasonably high accuracy, with $R^2 \approx 0.8$ for tortuosity and over 0.9 for permeability, while MAPE values remain within acceptable bounds.

However, for the most dense structures ($0.5\!<\!\varphi\!<\!0.6$), a substantial drop in performance is observed across all metrics. As the system approaches the percolation threshold, where critical phenomena begin to dominate transport behavior, the model's predictive capability breaks down. This highlights a fundamental limitation in the current approach—its reduced transferability in regimes exhibiting extreme porosity and critical connectivity transitions.

\modified{
The purpose of this experiment was to assess out-of-distribution generalization without modifying the training set, rather than to claim reliable performance across all porosity regimes. Low-porosity systems are particularly challenging because transport becomes increasingly controlled by narrow channels, connectivity effects, and proximity to the percolation threshold. The observed deterioration therefore identifies a limitation of the present model. Improving accuracy in this regime would likely require enriching the training data with additional low-porosity samples, for example through adaptive sampling or curriculum learning strategies.
}

\begin{figure}
    \centering
    \includegraphics[width=0.9\linewidth]{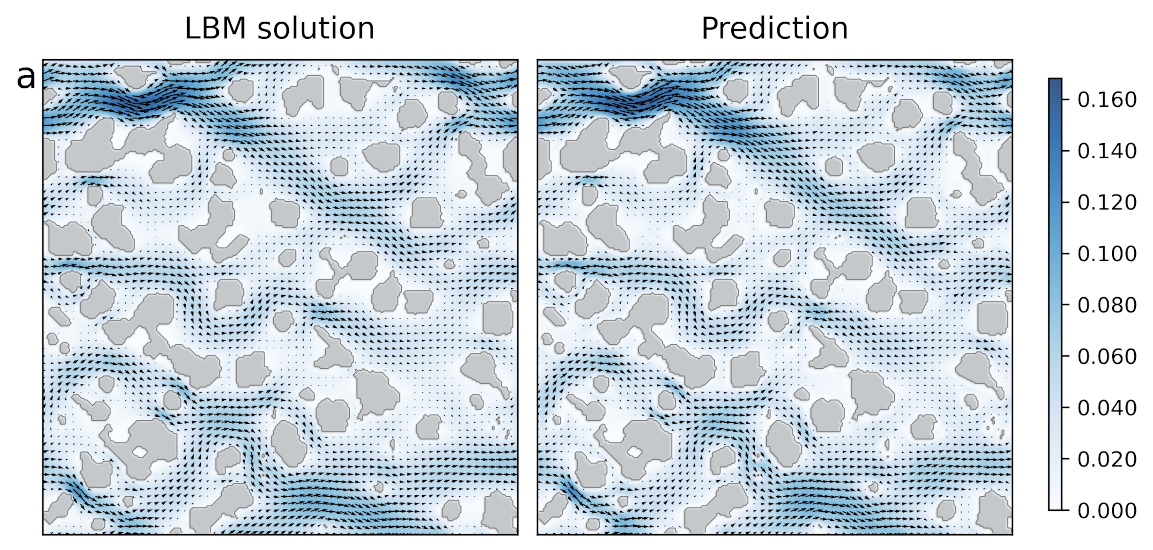}\\
    \includegraphics[width=0.9\linewidth]{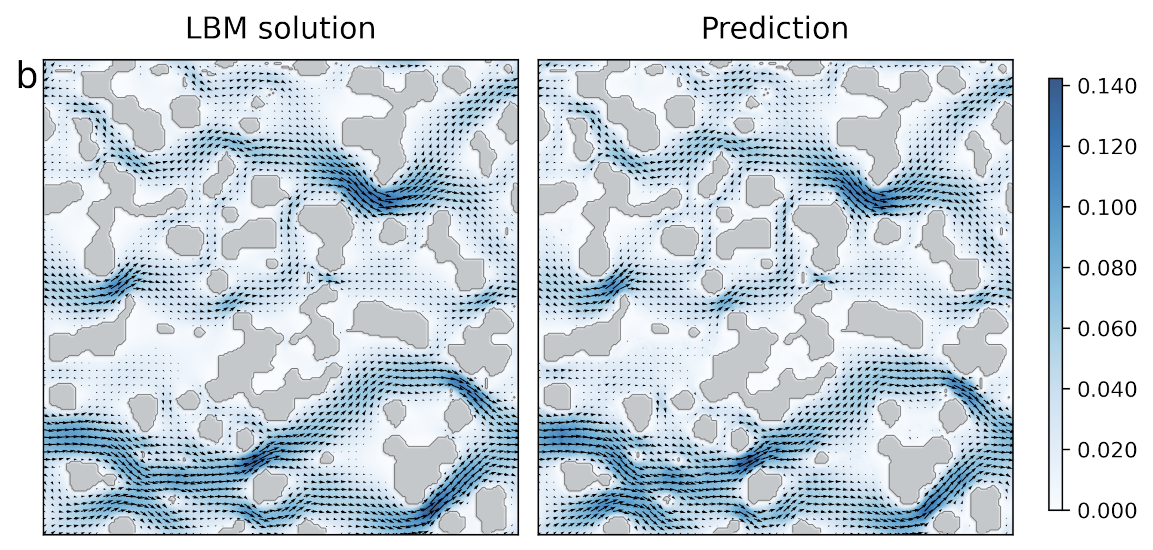}\\
    \includegraphics[width=0.9\linewidth]{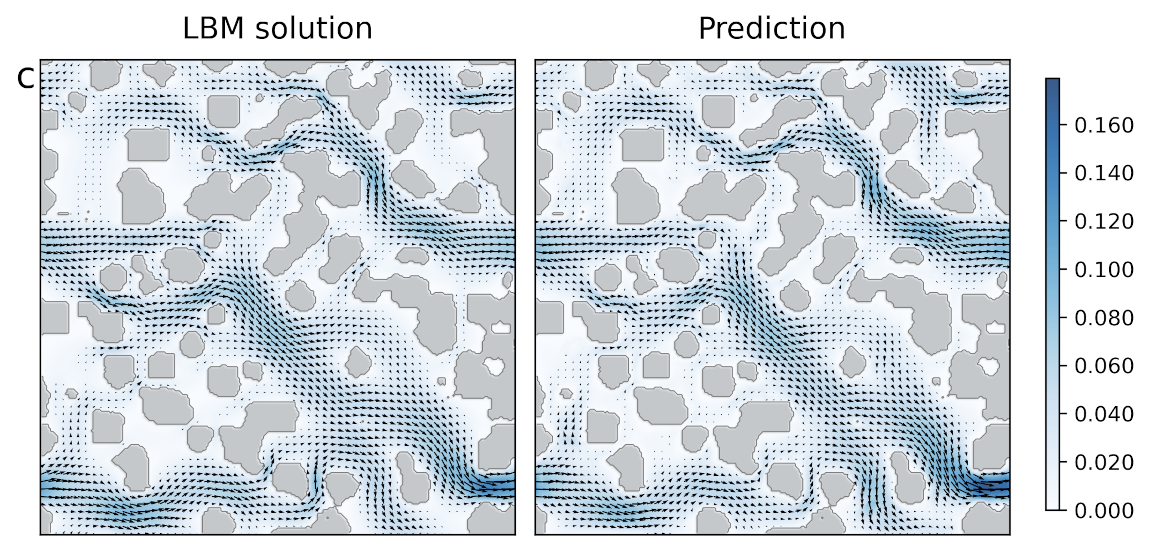}
    \caption{LBM computed and neural-network predicted velocities fields for exemplary real Li-O$_2$ electrodes. Systems have porosities: a) 0.728, b) 0.771, c) 0.802. }
    \label{fig:flow_lio2}
\end{figure}

\modified{
\subsection*{Real porous structures}
To evaluate the model on highly irregular real geometries, we tested it on the Li-O$_2$ electrodes dataset (DRP-1129v2 from the \cite{DPM_DRP1129, ProdanovicDPMP2025, Turhan2024}). This dataset consists of three-dimensional lithium-ion battery electrode microstructures reconstructed from Nano-CT scans, with porosity in the range of approximately 0.73–0.91. These structures differ substantially from the synthetic training data, exhibiting complex, heterogeneous pore networks formed through physical manufacturing processes. In our study, two-dimensional slices were extracted from the volumetric data and used as inputs, with corresponding velocity fields computed using LBM simulations.

The results, summarized in Table~\ref{tab:generalization}, show that the model maintains reasonable predictive accuracy despite the significant domain shift, with a velocity RMSE of $6.41 \cdot 10^{-3}$ and high agreement for permeability ($R^2 = 0.986$). At the same time, the accuracy for tortuosity decreases ($R^2 = 0.798$), reflecting the increased geometric complexity. These findings indicate that the model generalizes to realistic, highly irregular porous structures, while also highlighting the expected degradation in performance outside the training distribution. Figure \ref{fig:flow_lio2} shows representative LBM solutions and corresponding predictions made by the model.
}

\subsection{Use of CNN predictions for LBM initialization}

One of the problems with calculating flows in porous media using the LBM method at extremely high and low porosities is the hindered convergence of the method due to the diffusive nature of momentum transport \cite{Matyka08}. The classic, so-called 'cold' start of the LBM method involves running the calculation from a zero-velocity field. It turns out that even the imperfect velocity field predictions from neural networks can be an effective remedy to this problem and serve as a starting condition for the LBM method, which adds to the system at the end, smoothing out errors. 
\begin{figure}
    \centering
    \includegraphics[width=0.8\linewidth]{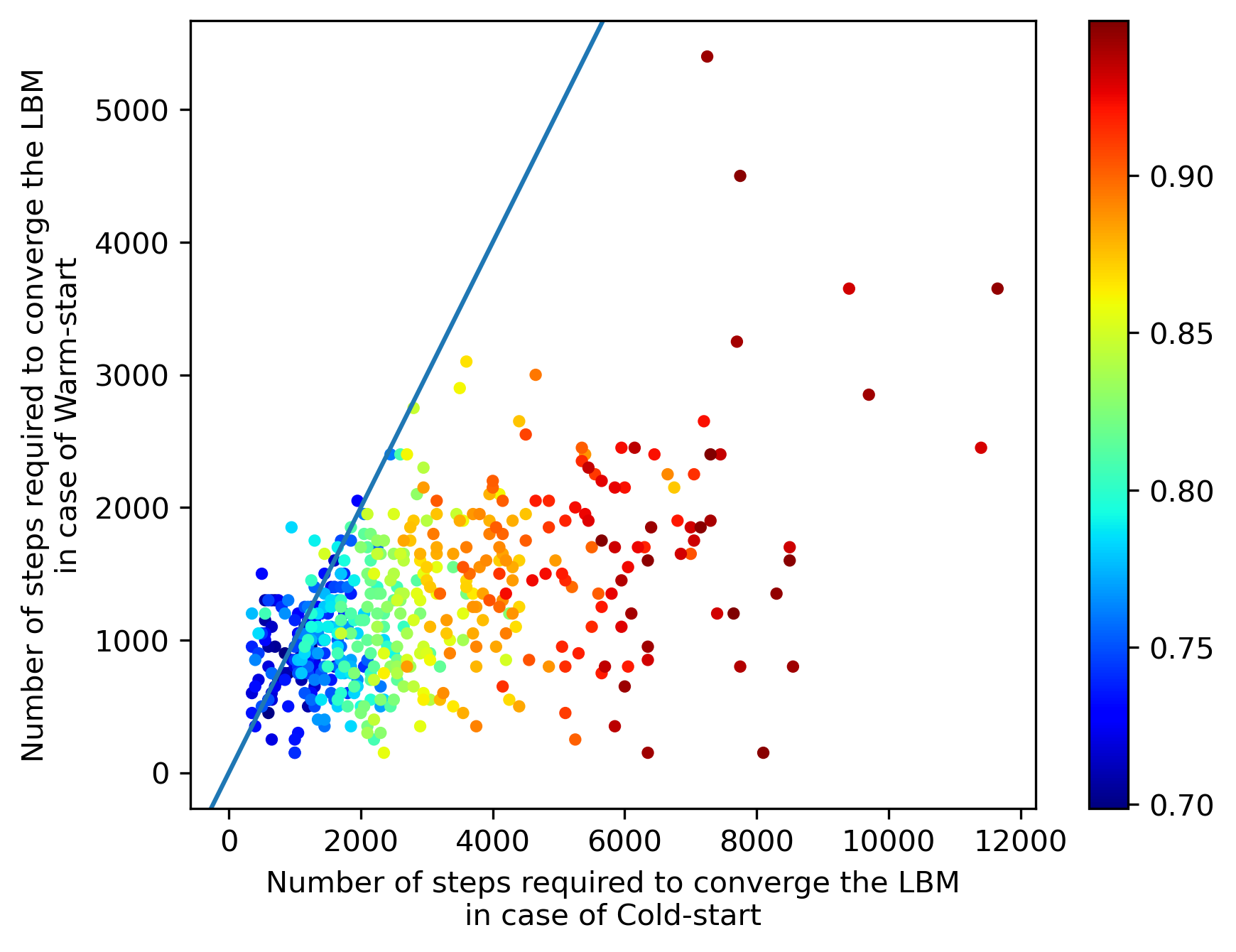}
    \caption{The graph presents the number of steps of the LBM when starting from the output of the CNN (so-called warm start, x axis) to the number of steps whether the LBM starts from zero (so-called cold start, y axis). All points below the line mean the LBM+CNN procedure gain time. Points are coloured by porosity. }
    \label{fig:speed}
\end{figure}
We decided to test this method on samples outside the scope of network learning and perform a test in which a neural network trained on trigonometric data makes predictions of flows in arrangements of overlapping figures, and then feeds such prepared data back to the LBM solver. 
As a result of starting from an approximate velocity field generated by the CNNs, the LBM method in $90\%$ systems converged faster to the correct solution (see Figure ~\ref{fig:speed}), compared to the solver that started from a zero velocity field. 
\modified{
The median reduction in the number of iterations was approximately $50\%$, meaning that in half of the tested systems the reduction was greater than $50\%$, and in half it was smaller. 
The interquartile range, given by $Q_1 \approx 22\%$ and $Q_3 \approx 68\%$, indicates that $50\%$ of the systems exhibit reductions within this interval. A bootstrap procedure (based on repeated resampling of the dataset and recomputation of the median) yields a $95\%$ confidence interval of approximately $[46\%, 54\%]$, which quantifies the statistical uncertainty of the estimated median reduction and shows that this value is stable across the dataset. A paired Wilcoxon signed-rank test confirms that this improvement is highly statistically significant ($p\approx 4 \times 10^{-70}$).
}

With the additional observation that the trained neural network produces results orders of magnitude faster, these findings provide hope for the practical application of our results.

\subsection{Computational time comparison}
Neural network enables fast inference. The average prediction time is just 5\,ms per structure on GPU (NVIDIA GeForce RTX 4090), while on a CPU (AMD Ryzen 9 5950X) it averages 190\,ms. Importantly, the inference time remains consistent regardless of porosity or other structural properties. In comparison, performing LBM calculations the same set of structures requires an average of 560\,ms per structure (utilizing the same CPU and the same number of cores). \modified{
It should be noted that this comparison concerns inference only. Training is computationally expensive but performed once and amortized over repeated predictions.
}

\section{Conclusions}
\label{secCON}
We have presented a complete pipeline for predicting pore-scale fluid flow through porous media using physics-informed convolutional neural networks. The main contribution of this work is the construction and evaluation of a custom loss function tailored to this problem, combining velocity-field reconstruction with penalties for flow inside solid regions, incompressibility, periodic consistency, and tortuosity matching.
Our analysis of weights for loss function components showed complex dependency of each part on the overall error of fluid flow predictions with the winning combination of $\alpha=5, \beta=1, \gamma=0.1$ and $\delta=0.01$. We suggest there is still space to optimize both the choice of the neural network architecture (ResNet-101 in our case), its size and combination of loss weights. \modified{
Future work may further improve performance through systematic architecture search, automated loss-weight optimization, or adaptive sampling of challenging regimes.
}

The key finding of this work, however, is generalization.
While the model is trained on a specific dataset of porous structures, the underlying architecture and training methodology are designed to be broadly applicable to a wide class of similar physical systems. The use of convolutional neural networks allows the model to learn spatially invariant features, enabling it to generalize to geometries and flow patterns that differ from those seen during training. Moreover, the inclusion of physical constraints in the loss function-such as incompressibility, zero-flow conditions inside solid regions, and periodicity-encourages the model to learn representations that are physically meaningful rather than overly dataset-specific. As a result, the trained model can be adapted to other porous systems, including those with different porosity distributions, obstacle shapes, or boundary conditions, provided that they share the same fundamental flow characteristics. For significantly different domains, transfer learning or fine-tuning on a small number of new samples may be sufficient to adapt the model while retaining much of the learned structure. Future work could explore this direction in more detail by systematically evaluating the model’s performance on out-of-distribution datasets or synthetic benchmarks with known analytical properties. 

\modified{
Despite the encouraging results, several limitations remain. The present implementation is restricted to two-dimensional, fixed-resolution geometries, and its accuracy decreases for strongly out-of-distribution cases, such as low-porosity structures near connectivity transitions. However, the approach is, by design, naturally extendable to three-dimensional systems: the same encoder--decoder concept, backbone-based architecture, and physics-informed loss strategy can be retained, with two-dimensional convolutional operations replaced by their three-dimensional counterparts. Moreover, the physical constraints are imposed through penalty terms and are therefore only approximately enforced. Future work should explore full three-dimensional implementations, transfer learning for experimentally obtained microstructures, systematic optimization of loss weights, and possible integration with other physics-informed or operator-learning frameworks, such as PINNs, Fourier Neural Operators, or DeepONets.
}

\section{Code and data availability}
Both the dataset and the code are publicly available in a dedicated \href{https://github.com/dioscuri-tda/PhysInfPorousFluidFlow}{GitHub repository}. Moreover, we provide the best-performing pretrained model, which can be used directly without requiring retraining.

\section*{Author Contributions}
R.T., P.D. and M.M. started and designed the project. R.T. developed codebase for neural networks, train the models and analyzed the results M.M. prepared fluid flow simulations codes. R.T. and M.M. designed the loss function to incorporate flow conditions into training. P.D. suggested time analysis and LBM initialization procedure. R.T. and M.M. wrote the first draft of manuscript. All authors participated in writing the text and discussing the results.

\section{Funding}
R. Topolnicki and P. Dłotko acknowledges the support of Dioscuri program initiated by the Max Planck Society, jointly managed with the National Science Centre (Poland), and mutually funded by the Polish Ministry of Science and Higher Education and the German Federal Ministry of Education
and Research. Funded by National Science Centre, Poland under the OPUS call in the Weave programme 2021/43/I/ST3/00228. This research was funded in whole or in part by National Science Centre (2021/43/I/ST3/00228). For the purpose of Open Access, the author has applied a CC-BY public copyright licence to any Author Accepted Manuscript (AAM) version arising from this submission. 

M. Matyka acknowledges the financial support from the Slovenian Research And Innovation Agency (ARIS) research core funding No.\ P2-0095.

Authors acknowledge the financial support from M-ERA.NET’s PORMETALOMICS project supported by 
\newline MCIN/AEI/10.13039/501100011033 and the European Union’s NextGenerationEU/PRTR funds. Financial support from the PORMETALOMICS project, funded by the National Science Centre, Poland (project no. 2021/03/Y/ST5/00232) within the M-ERA.NET 3 programme, is also gratefully acknowledged. This project has received funding from the European Union’s Horizon 2020 research and innovation programme under grant agreement No 958174.

\section*{Acknowledgements}

The computations in this work were partially supported by the Center for Artificial Intelligence at Adam Mickiewicz University. We would like to thank Prof. K. Jassem and Dr. B. Naskrecki for granting us access to the Center’s infrastructure.
The calculations were made with the support of the Interdisciplinary Center for Mathematical and Computational Modeling of the University of Warsaw (ICM UW) under the computational grant no g105-2705.

\bibliographystyle{elsarticle-harv} 
%\bibliography{cnn}

\clearpage

% Supplementary Information formatting
\onecolumn
\setstretch{1.15}

\appendix
\renewcommand{\thefigure}{S\arabic{figure}}
\renewcommand{\thetable}{S\arabic{table}}
\renewcommand{\theequation}{S\arabic{equation}}
\renewcommand{\thesection}{S\arabic{section}}

\setcounter{figure}{0}
\setcounter{table}{0}
\setcounter{equation}{0}
\setcounter{section}{0}

\section*{Supplementary Information}
\addcontentsline{toc}{section}{Supplementary Information}
\begin{center}
\Huge{Supplementary Information} 
\vspace{2cm} \\
\LARGE{Physics-informed convolutional neural networks for fluid flow through porous media}
\end{center}

\vspace{2cm} 
\tableofcontents
\newpage

\section{Error Bounds}
\begin{figure}[h!]
    \centering
    \includegraphics[width=0.6\linewidth]{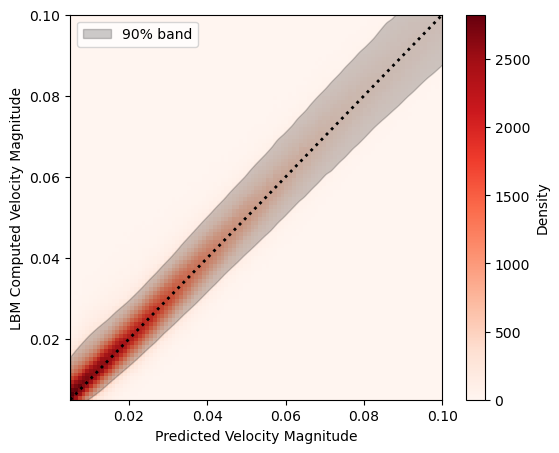}
    \caption{
    Two-dimensional histogram of ground-truth (LBM-computed) velocity magnitude $|v|_{\mathrm{LBM}}$ versus predicted velocity magnitude $|v|_{\mathrm{pred}}$. The color scale represents the density of samples. The dashed line indicates the ideal relation $|v|_{\mathrm{LBM}} = |v|_{\mathrm{pred}}$, while the shaded region denotes the estimated $90\%$ conditional confidence band for $|v|_{\mathrm{LBM}}$ given $|v|_{\mathrm{pred}}$, obtained from empirical quantiles of the residuals $r = |v|_{\mathrm{LBM}} - |v|_{\mathrm{pred}}$. The narrow band in the high-density region indicates strong agreement between predictions and ground truth for moderate velocities, whereas the widening of the band at higher velocities reflects increased uncertainty in more challenging flow regimes.}
    \label{fig:vel_histogram}
\end{figure}

To assess the robustness of the surrogate in the regimes highlighted by the referee, we complement global error metrics with a conditional uncertainty analysis based on the velocity magnitude $|v|$. 
In Fig.~\ref{fig:vel_histogram}, we plot a two-dimensional histogram of $|v|_{\mathrm{pred}}$ versus $|v|_{\mathrm{LBM}}$, which shows both the concentration of samples around the ideal relation $y=x$ and the spread of the prediction error as a function of the predicted velocity magnitude. 
The purpose of estimating confidence bounds is to move beyond a single averaged score and quantify how predictive uncertainty changes across the state space, in particular in difficult low-porosity regimes where transport pathways become constricted and the flow becomes more heterogeneous. 
To this end, we define residuals
\begin{equation}
r = |v|_{\mathrm{LBM}} - |v|_{\mathrm{pred}},
\end{equation}
and estimate conditional lower and upper quantiles of $r$ within bins of the predicted velocity magnitude $|v|_{\mathrm{pred}}$. Denoting the conditional quantile function by $Q_{\alpha}(\cdot)$, the empirical estimates correspond to
\begin{equation}
q_{\alpha}(x) = Q_{\alpha}\big(r \,\big|\, |v|_{\mathrm{pred}} = x \big),
\end{equation}
where $\alpha \in \{0.05, 0.95\}$ in the present work. These quantiles define lower and upper bounds for the ground-truth velocity magnitude conditioned on the prediction:
\begin{equation}
\underline{v}(x) = x + q_{0.05}(x), \qquad
\overline{v}(x) = x + q_{0.95}(x).
\end{equation}

The discrete estimates $\{(x_i, q_{\alpha}(x_i))\}$ obtained from binning are then converted into continuous functions using a shape-preserving piecewise cubic Hermite interpolant (PCHIP), \cite{FritschCarlson1980,FritschButland1984}. In practice, this corresponds to constructing a piecewise cubic function $s_{\alpha}(x)$ such that
\begin{equation}
s_{\alpha}(x_i) = q_{\alpha}(x_i), \quad i = 1,\dots,N,
\end{equation}
with continuity of both $s_{\alpha}(x)$ and its first derivative, while preserving the local monotonicity of the data. The final confidence bounds are therefore given by
\begin{equation}
\underline{v}(x) = x + s_{0.05}(x), \qquad
\overline{v}(x) = x + s_{0.95}(x).
\end{equation}

These discrete quantile estimates are then converted into continuous lower and upper bound functions using a shape-preserving piecewise cubic Hermite interpolant, PCHIP, \cite{FritschCarlson1980,FritschButland1984}. In practice, this interpolation constructs a cubic polynomial within each bin interval while enforcing continuity of both the function and its first derivative across bin boundaries. Unlike standard cubic splines, which may introduce non-physical oscillations or overshoots in regions with sparse or rapidly changing data, PCHIP explicitly preserves the local monotonicity and shape of the underlying quantile estimates. As a result, the interpolated bounds remain consistent with the empirical trends observed in the binned data, providing smooth yet faithful approximations of the conditional quantile curves without introducing artificial extrema or distortions.

\begin{figure}[h!]
    \centering
    a) \includegraphics[width=0.9\linewidth]{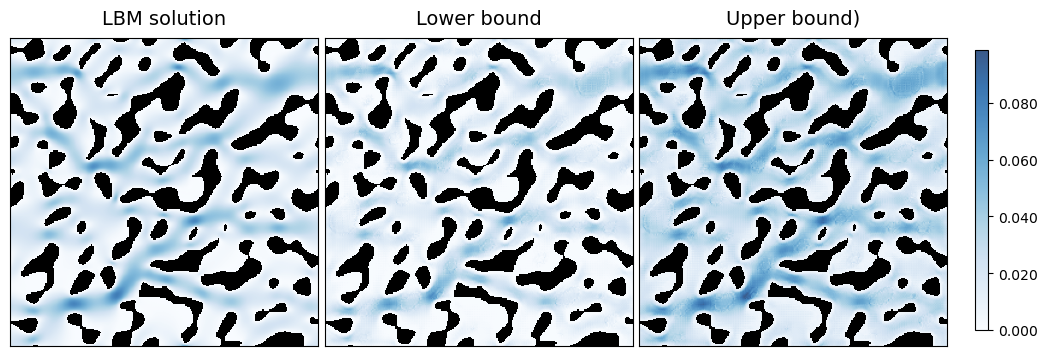} \\
    b) \includegraphics[width=0.9\linewidth]{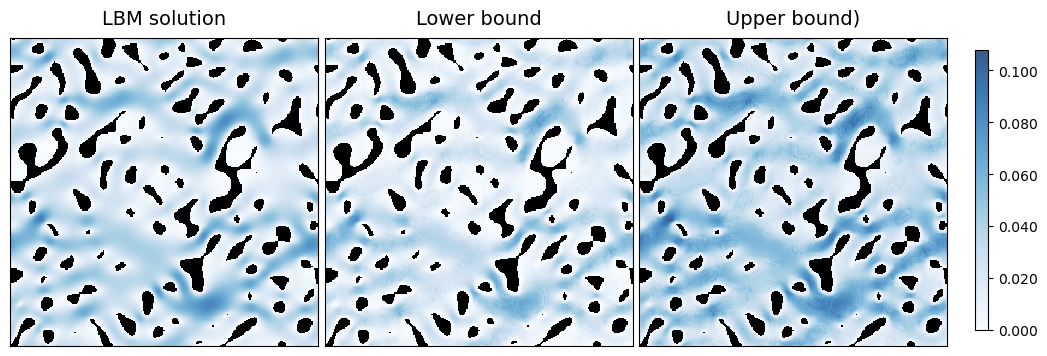} \\
    c) \includegraphics[width=0.9\linewidth]{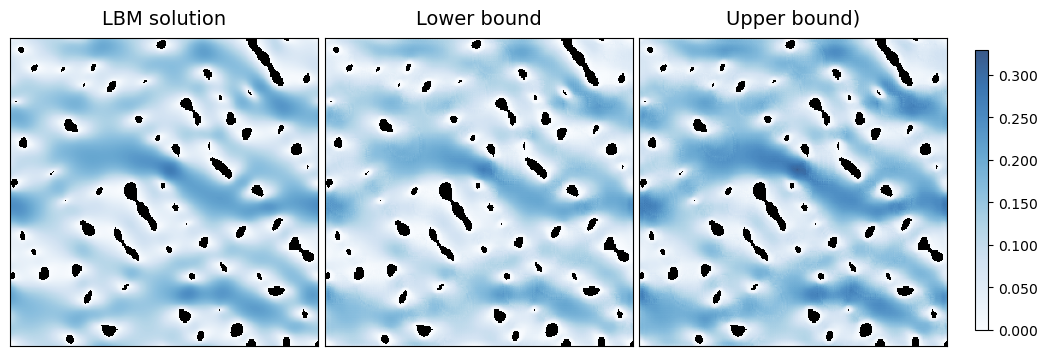} 
    \caption{
    Spatial visualization of conditional confidence bounds for the velocity magnitude $|v|$. The left panel shows the ground-truth LBM solution, while the central and right panels display the estimated lower and upper confidence bounds, respectively, obtained from the conditional quantile model. Solid regions correspond to the pore structure. The bounds are evaluated pointwise as functions of the predicted velocity magnitude $|v|_{\mathrm{pred}}$, resulting in spatially varying uncertainty estimates. 
    Systems have porosities: a) 0.713, b) 0.817, c) 0.931.}
    \label{fig:vel_bounds_panels}
\end{figure}

The resulting bounds are shown globally as a band around the diagonal in Fig.~\ref{fig:vel_histogram}, and spatially in Fig.~\ref{fig:vel_bounds_panels}, where the lower and upper confidence limits are evaluated pointwise as functions of the predicted velocity magnitude $|v|_{\mathrm{pred}}$. The multi-panel visualization provides additional insight into the spatial structure of the uncertainty: regions of high velocity, typically corresponding to preferential flow channels, remain tightly constrained, whereas low-velocity regions near solid boundaries and in poorly connected pore spaces exhibit a visibly wider confidence envelope. These regions are associated with reduced porosity and proximity to the percolation threshold, where the flow becomes more sensitive to small geometric variations.
The predicted fields remain consistent with the topology and spatial organization of the LBM solution, and the ground-truth solution is, in most regions, contained within the estimated bounds. This indicates that the model retains meaningful predictive capability while appropriately reflecting increased uncertainty in more complex flow configurations. 

Overall, this conditional quantile-based analysis provides a more detailed characterization of model performance, highlighting how predictive reliability varies across the domain and offering spatially resolved bounds on the expected prediction error, rather than relying solely on global summary metrics.

The same uncertainty quantification framework can be applied independently to the individual components of the velocity vector. In this case, the analysis is performed for each component (e.g., $v_x$ and $v_y$) by constructing conditional confidence intervals for the ground-truth component given its predicted value. This involves defining residuals $r = v_{\mathrm{LBM}} - v_{\mathrm{pred}}$ for each component and estimating conditional quantiles of $r$ within bins of the predicted component. This extension allows for component-wise uncertainty characterization, capturing potential anisotropies in the prediction error while preserving the local structure of the flow field. In practice, the results are qualitatively consistent with the magnitude-based analysis, while providing additional directional insight into the model performance.

\section{Training procedure details}
Training was performed on a workstation equipped with an NVIDIA GeForce RTX 4090 GPU with 24 GB of VRAM and an AMD Ryzen 9 5950X CPU. The dataset was randomly split into training, validation, and test subsets containing 70\%, 15\%, and 15\% of the samples, respectively. A batch size of 32 was used in all experiments. During training, the loss was monitored on the validation set, and early stopping was applied with a patience of 20 epochs. The total number of epochs varied between architectures, but training typically required approximately 250 epochs. Optimization was performed using the Adam algorithm with an initial learning rate of $5\times10^{-4}$. The learning rate was updated using a step scheduler with step size 10 and decay factor \texttt{gamma=0.8}.

We found that the most stable convergence was obtained using a two-stage training procedure. In the first stage, the model was trained with the obstacle loss enabled and the remaining physics-based terms disabled, i.e. with $\alpha=5$ and $\beta=\gamma=\delta=0$. In the second stage, training was resumed from the weights obtained in the first stage using the full loss function with the final hyperparameter values $\alpha=5, \beta=1, \gamma=0.1$ and $\delta=0.01$.

\section{Loss function hyperparameters}
The loss function introduced in Eq. (11) of the manuscript contains four hyperparameters: $\alpha$, $\beta$, $\gamma$, and $\delta$, which control the relative importance of different physical constraints. The terms weighted by $\alpha$ and $\beta$ act locally, enforcing agreement with the reference velocity field and local physical constraints such as zero velocity inside obstacles and divergence-free flow. In contrast, the terms weighted by $\gamma$ and $\delta$ operate at a global level, enforcing periodic boundary conditions and consistency of derived macroscopic quantities, such as tortuosity.

These parameters can therefore be interpreted as balancing local fidelity against global physical consistency. While their values are selected through standard hyperparameter tuning, they also provide a degree of flexibility, allowing the user to emphasize specific physical properties depending on the application.

To further investigate their influence, we performed additional experiments in which each hyperparameter was varied over approximately two orders of magnitude, while the others were kept fixed. The tested values were:
$\alpha={1, 5, 10}$, $\beta={0.1, 1, 10}$, $\gamma={0.01, 0.1, 1}$, and $\delta={0.001, 0.01, 0.1}$, where the middle values correspond to those used in the main manuscript. Each experiment was repeated three times with different data splits, and median values are reported.

Table~\ref{tab:sup_losscomplexity} summarizes the results. The table is organized in blocks corresponding to the parameter being varied, with the associated loss term highlighted for clarity. Increasing a given hyperparameter consistently reduces the corresponding component of the loss function, as expected. However, this does not necessarily translate into a reduction of the overall mean squared error (MSE), indicating non-trivial interactions between the different loss terms.

In many cases, adding additional loss components improves the overall prediction accuracy, which may be attributed to improved optimization and avoidance of local minima. Notably, the parameter set used in the main manuscript ($\alpha=5$, $\beta=1$, $\gamma=0.1$, $\delta=0.01$) corresponds to one of the lowest MSE values observed.

The interplay between local and global terms is not trivial. For example, introducing the global periodicity term ($\gamma > 0$) reduces the MSE, even though it primarily enforces a global constraint. Adding the tortuosity term ($\delta > 0$) slightly increases local error metrics but significantly improves the accuracy of predicted macroscopic quantities.

It is worth noting that the same set of hyperparameters, i.e. $\alpha=5$, $\beta=1$, $\gamma=0.1$, and $\delta=0.01$, was also used in the Fourier Neural Operator experiments described in Section~\ref{sec:fno}. A similar trend is observed in that case: the inclusion of additional loss terms leads to a consistent improvement in all evaluated metrics. This indicates that the beneficial effect of the proposed loss formulation is not specific to convolutional architectures, but extends to operator-learning approaches as well.

Overall, these results indicate that the loss function components interact in a coupled manner, and that the choice of hyperparameters reflects a trade-off between different physical properties. The selected parameter values represent a balanced compromise between local accuracy and global physical consistency.

\begin{table}[]
\centering
\caption{
Performance metrics for models trained with different values of loss-function hyperparameters. All models use the same ResNet-101 architecture. Median values over three independent runs with different train/validation/test splits are reported. The table illustrates how individual loss components and prediction accuracy respond to changes in the weighting coefficients.
}
\label{tab:sup_losscomplexity}
\begin{tabular}{rrrr|rrrrrr}
\multicolumn{1}{l}{$\alpha$} & 
\multicolumn{1}{l}{$\beta$} & 
\multicolumn{1}{l}{$\gamma$} & 
\multicolumn{1}{l}{$\delta$} & 
\multicolumn{1}{l}{$\mathcal{L}_\text{vel}$} & 
\multicolumn{1}{l}{$\mathcal{L}_\text{inside}$} & 
\multicolumn{1}{l}{$\mathcal{L}_\text{div}$} & 
\multicolumn{1}{l}{$\mathcal{L}_\text{periodic}$} & 
\multicolumn{1}{l}{$\mathcal{L}_\text{tort}$} & 
\multicolumn{1}{l}{$R^2_\tau$} \\ \hline

{\color[HTML]{0000FF} 1} & 0 & 0 & 0 & 8.79E-05 & {\color[HTML]{0000FF} 1.58E-05} & 4.15E-06 & 1.04E-04 & 1.73E-03 & 0.757 \\
{\color[HTML]{0000FF} 5} & 0 & 0 & 0 & 3.68E-05 & {\color[HTML]{0000FF} 4.43E-06} & 6.41E-06 & 3.34E-05 & 6.85E-04 & 0.901 \\
{\color[HTML]{0000FF} 10} & 0 & 0 & 0 & 6.61E-05 & {\color[HTML]{0000FF} 2.60E-06} & 1.28E-05 & 6.26E-05 & 2.86E-03 & 0.581 \\ \hline

5 & {\color[HTML]{38761D} 0.1} & 0 & 0 & 2.14E-05 & 1.40E-06 & {\color[HTML]{38761D} 3.53E-06} & 2.32E-05 & 9.23E-04 & 0.859 \\
5 & {\color[HTML]{38761D} 1} & 0 & 0 & 2.00E-05 & 1.45E-06 & {\color[HTML]{38761D} 2.00E-06} & 2.30E-05 & 1.09E-03 & 0.833 \\
5 & {\color[HTML]{38761D} 10} & 0 & 0 & 2.08E-05 & 1.70E-06 & {\color[HTML]{38761D} 7.51E-07} & 2.16E-05 & 1.52E-03 & 0.768 \\ \hline

5 & 1 & {\color[HTML]{CC0000} 0.01} & 0 & 2.02E-05 & 1.56E-06 & 1.90E-06 & {\color[HTML]{CC0000} 2.25E-05} & 7.12E-04 & 0.891 \\
5 & 1 & {\color[HTML]{CC0000} 0.1} & 0 & 1.79E-05 & 1.42E-06 & 1.84E-06 & {\color[HTML]{CC0000} 1.86E-05} & 8.71E-04 & 0.873 \\
5 & 1 & {\color[HTML]{CC0000} 1} & 0 & 1.71E-05 & 1.57E-06 & 1.80E-06 & {\color[HTML]{CC0000} 9.84E-06} & 1.52E-03 & 0.787 \\ \hline

5 & 1 & 0.1 & {\color[HTML]{E69138} 0.001} & 2.52E-05 & 2.26E-06 & 3.39E-06 & 2.50E-05 & {\color[HTML]{E69138} 3.92E-04} & {\color[HTML]{E69138} 0.943} \\
5 & 1 & 0.1 & {\color[HTML]{E69138} 0.01} & 2.07E-05 & 1.85E-06 & 2.09E-06 & 2.22E-05 & {\color[HTML]{E69138} 1.24E-04} & {\color[HTML]{E69138} 0.983} \\
5 & 1 & 0.1 & {\color[HTML]{E69138} 0.1} & 2.67E-05 & 2.15E-06 & 2.59E-06 & 2.63E-05 & {\color[HTML]{E69138} 9.58E-05} & {\color[HTML]{E69138} 0.986} \\
\end{tabular}
\end{table}

\section{Fourier Neural Operators}
\label{sec:fno}
\begin{table*}[]
\centering
\caption{
Comparison between the best-performing CNN architecture (ResNet101, see Table 2 in the manuscript) and the Fourier Neural Operator (FNO) models. Two FNO variants are considered: one trained with a simplified loss function and one trained using the full loss formulation adopted for CNNs. In all FNO cases, $\alpha=5$.}
\begin{tabular}{l|r|rrr|rrr}
 & \multicolumn{1}{c|}{velocity} & \multicolumn{3}{c|}{tortuosity}& \multicolumn{3}{c}{permeability} \\
 & {RMSE} & {RMSE} & {MAPE} & {$R^2$} & {RMSE} & {MAPE} & {$R^2$} \\ \hline
ResNet101 & 5.10E-03 & 1.13E-02 & 0.6 & 0.983 & 8.69E+02 & 4.8 & 0.997  \\
FNO $\beta=0, \gamma=0, \delta=0$ & 5.52E-03 & 2.71E-02 & 1.37 & 0.897 & 1.26E+03 & 5.49 & 0.993 \\
FNO $\beta=1, \gamma=0.1, \delta=0.01$ & 5.48E-03 & 2.27E-02 & 0.98 & 0.928 & 1.31E+03 & 5.85 & 0.992 \\
\end{tabular}
\label{tab:fno_results}
\end{table*}

In addition to convolutional neural networks, we explored the use of Fourier Neural Operators (FNOs) as an alternative model class. FNOs are a recently proposed class of neural operators designed to learn mappings between function spaces, rather than between discrete images \cite{li2021fourier, kovachki2023neural}. They achieve this by performing global convolutions in the Fourier domain, enabling efficient modeling of long-range spatial dependencies. This makes them particularly appealing for problems governed by partial differential equations, such as fluid flow in porous media.

In our implementation, the FNO model operates on regular grids and consists of a sequence of spectral convolution layers interleaved with pointwise transformations. The input is augmented with spatial coordinate information, allowing the network to learn position-dependent mappings. The model was trained to predict the two components of the velocity field directly, using the same dataset and train–validation split as for the CNN-based models. After preliminary exploration of model capacity, we selected a configuration with 16 retained Fourier modes per spatial dimension, a channel width of 256, and 4 stacked spectral layers, which provided the best performance among the tested variants.

Two versions of the FNO model were trained. The first one employed a simplified loss function, corresponding to Eqs.~(11–13) in the manuscript, with $\alpha=5$ and $\beta=\gamma=\delta=0$. The second version used the full loss formulation introduced for the CNN models, with $\beta=1$, $\gamma=0.1$, and $\delta=0.01$. This allows us to assess whether the additional physics-informed terms in the loss function provide consistent benefits across different model architectures.

The results are summarized in Table~\ref{tab:fno_results}. The FNO model achieves predictive performance of the same order of magnitude as the CNN-based approach, particularly for the velocity field reconstruction. More importantly, the inclusion of the additional loss terms leads to a systematic improvement across all evaluated metrics. In particular, the tortuosity prediction error decreases and the coefficient of determination $R^2$ increases when the full loss formulation is used. A similar trend is observed for permeability.

These observations indicate that the proposed loss function contributes to improved model performance independently of the underlying architecture. In particular, the gains obtained by incorporating the additional loss terms are not specific to convolutional networks, but extend to operator-learning approaches such as FNO. This suggests that the loss formulation plays a key role in guiding the model towards physically consistent predictions.

\bibliographystyle{elsarticle-harv}

\end{document}